\documentclass[twoside,english,format=manuscript,screen=true]{acmart}
\usepackage[T1]{fontenc}
\usepackage[latin9]{inputenc}
\pdfoutput=1
\usepackage{color}
\usepackage{array}
\usepackage{verbatim}
\usepackage{multirow}
\usepackage{graphicx}

\makeatletter

\providecommand{\tabularnewline}{\\}

\usepackage{booktabs} 

\usepackage[ruled]{algorithm2e} 

\SetAlFnt{\small}
\SetAlCapFnt{\small}
\SetAlCapNameFnt{\small}
\SetAlCapHSkip{0pt}
\IncMargin{-\parindent}

\makeatother

\usepackage{babel}
\begin{document}
\acmYear{2019} 
\copyrightyear{2019} 

\setcopyright{acmcopyright} 

\title{Multifaceted Analysis of Fine-Tuning in Deep Model for Visual
Recognition}

\author{Xiangyang Li} 
\affiliation{ 
 \institution{Institute of Computing Technology, CAS}  
}
\affiliation{ 
 \institution{University of Chinese Academy of Sciences}
}
\author{Luis Herranz} 
\affiliation{%
 \institution{Computer Vision Center, Universitat Aut{\`o}noma de Barcelona }  
}
\author{Shuqiang Jiang}
\affiliation{ 
 \institution{Institute of Computing Technology, CAS}  
}
\affiliation{ 
 \institution{University of Chinese Academy of Sciences}
}
\authorsaddresses{
Authors' addresses: X. Li and S. Jiang, Key Laboratory of Intelligent Information Processing of Chinese Academy of Sciences (CAS), Institute of Computing Technology, CAS, Beijing 100190, China; and University of Chinese Academy of Sciences, Beijing 100049, China; emails: xiangyang.li@vipl.ict.ac.cn, sqjiang@ict.ac.cn; L. Herranz, Learning and Machine Perception (LAMP) group of the Computer Vision Centre,  Universitat Aut{\`o}noma de Barcelona, 08193 Bellaterra, Barcelona, Spain; email: lherranz@cvc.uab.es (This work was performed when L. Herranz was at ICT, CAS)}

\begin{abstract}

\textcolor{black}{In recent years, convolutional neural networks (CNNs)
have achieved impressive performance for various visual recognition
scenarios. CNNs trained on large labeled datasets can not only obtain
significant performance on most challenging benchmarks but also provide
powerful representations, which can be used to a wide range of other
tasks. However, the requirement of massive amounts of data to train
deep neural networks is a major drawback of these models, as the data
available is usually limited or imbalanced. Fine-tuning (FT) is an
effective way to transfer knowledge learned in a source dataset to
a target task. In this paper, we introduce and systematically investigate
several factors that influence the performance of fine-tuning for
visual recognition. These factors include parameters for the retraining
procedure (e.g., the initial learning rate of fine-tuning), the distribution
of the source and target data (e.g., the number of categories in the
source dataset, the distance between the source and target datasets)
and so on. We quantitatively and qualitatively analyze these factors,
evaluate their influence, and present many empirical observations.
The results reveal insights into what fine-tuning changes CNN parameters
and provide useful and evidence-backed intuitions about how to implement
fine-tuning for computer vision tasks.}

\end{abstract}

%
%


\maketitle


\section{Introduction}

Visual recognition is a fundamental concern of computer vision in
the big data age. Over the past years, it has achieved significant
progress due to the rapid development of ubiquitous sensing technologies
of collecting image data and the available of large computational
resources to train big models. One of the huge successes is deep convolutional
neural networks (CNNs), which have achieved excellent performance
on large number of visual tasks, such as recognition~\cite{CNNs,He2014,NIN,VERY_DEEP,GIONG_DEEPER,BN_2015,2016Residual,DenseConvnet_2016,I_Map_2016,IncepRes2017,SEN_2017},
object detection\ \cite{OVER_FEAT_2014,RICH_FEATURE,Fast_RCNN,Fsater-RCNN},
segmentation\ \cite{Full_Segmentation_2015,Segmentation_2016} and
so on. These models are mainly built upon the convolution operation,
which extracts discriminative and informative features by gradually
integrating spatial and channel information within local receptive
fields. In this manner, they learn visual representations layer-by-layer,
where low features (e.g., color blobs, edges and simple shapes, which
are applicable to many datasets and tasks) are general and high layer
features are specific, which depend much on the target dataset and
task\ \cite{TRANS_FT}. 

\textcolor{black}{Krizhevsky \textit{et al.}~\cite{CNNs} firstly
train }\textcolor{black}{\emph{AlexNet}}\textcolor{black}{\ \cite{CNNs}
on ImageNet for the large scale visual recognition challenge in 2012
(ILVRC 2012)\ \cite{ImageNet-2015-ijcv}. Their successes are that
they can not only obtain significant performance on most challenging
datasets\ \cite{ImageNet-2015-ijcv,Places_2017}, but also provide
powerful representations which can be used to other tasks or different
datasets\ \cite{DECAF_2014,OFF_SHELF_2014,DEVIL,OEDERLESS_POOLING_2014,PYRAMID_P_2015S}.
After this, CNNs have been successfully applied to numerous visual
tasks. The approaches employing CNNs can be divided into two categories
according to whether having abundant data to train deep neural networks.}

\begin{figure}
\noindent\begin{minipage}[t]{1\columnwidth}%
\begin{center}
\includegraphics[width=0.42\columnwidth]{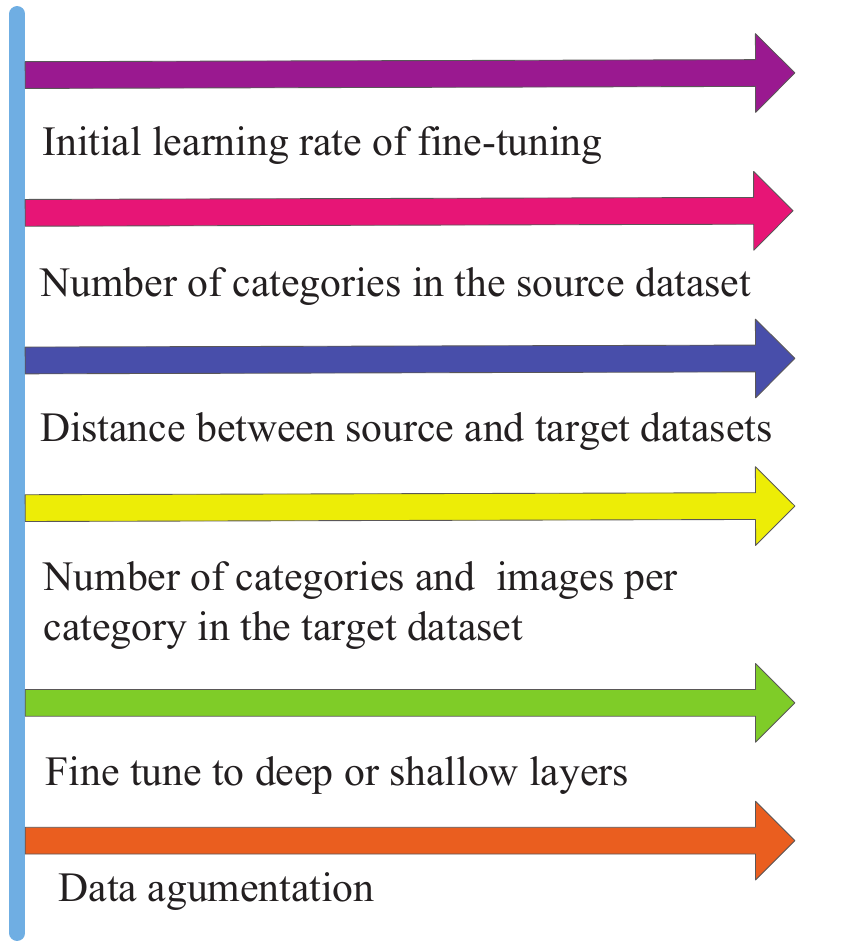}
\par\end{center}%
\end{minipage}

\caption{\textcolor{black}{Factors that influence the performance of fine-tuning.}\label{fig:Factors-that-influence}}
\end{figure}

\textcolor{black}{With the available of large annotated target datasets
such as ImageNet\ \cite{ImageNet-2015-ijcv}, Places\ \cite{Places_2017},
the first way is to directly train a model from scratch on the target
dataset. After the success of }\textcolor{black}{\emph{AlexNet}}\textcolor{black}{\ \cite{CNNs},
many complex architectures have been proposed such as }\textcolor{black}{\emph{VGGNet}}\textcolor{black}{\ \cite{VERY_DEEP},
}\textcolor{black}{\emph{GoogLeNet}}\textcolor{black}{\ \cite{GIONG_DEEPER},
}\textcolor{black}{\emph{ResNet}}\textcolor{black}{\ \cite{2016Residual},
}\textcolor{black}{\emph{SENet}}\textcolor{black}{\ \cite{SEN_2017}
and so on. These high-capacity models have brought excellent progress
by mainly increasing network depth, width, as well as enhancing connection
between different layers and channels of feature maps. For example,
Symonyan \textit{et al.}~\cite{VERY_DEEP} investigate the effect
of network depth with the recognition accuracy and propose a 19-layer
architecture which outperforms previous neural networks with a huge
margin. Hu\ \textit{et al.}~\cite{SEN_2017} focus on the channel
interdependence in feature maps and propose}\textcolor{black}{\emph{
SENet}}\textcolor{black}{~\cite{SEN_2017}}\textcolor{black}{\emph{
}}\textcolor{black}{which has obtained the best performance in ILSVRC
2017. Luan\ \textit{et al.}\ \cite{Gabor-CN-TIP-2018} propose
Gabor Convolutional Networks (GCNs) which incorporate gabor filters
to CNNs to enhance deep feature representations with steerable orientation
and scale capacities. Wu\ \textit{et al.}\ \cite{Joi_Cross-TIE}\ propose
a model to learn unified hash codes as well as deep hash functions.
Liu\ \textit{et al.}\ \cite{Single-image_super_2019} design a
multi-scale architecture based on the network simulation of wavelet
multi-resolution analysis. Wu\ \textit{et al.}\ \cite{U-Hasing-retrieval-2018}
introduce a deep model which integrates deep convolutional networks
with binary latent representation learning. Li\ \textit{et al.}\ \cite{IIP-2014}\ combine
CNN and recursive neural network (RNN) for visual recognition. Wu\ \textit{et al.}\ \cite{Video-TIP2019}\ propose
a model where feature learning and hash function learning are jointly
integrated. }

However, for many object recognition or scene classification datasets,
there are always not so many available labeled images to feed such
big models. For these situations, the second way is to just employ
a pre-trained CNN as a feature extractor. Approaches of this category
firstly take the whole image as the input for a pre-trained CNN to
extract high layers activations as image representations and then
utilize them to train simple classifiers to obtain recognition results.
The combination of deep representations and a simple classifier has
been a preferred solution for many visual recognition tasks~\cite{DECAF_2014,DEVIL,OEDERLESS_POOLING_2014,OFF_SHELF_2014,PYRAMID_P_2015S,CROSS_CONV_LAYERS}.
For example, Donahue \textit{et al.} \cite{DECAF_2014} suggest that
the deep CNN trained on ImageNet is an effective feature extractor
and provide large evidence to support this claim. Gong \textit{et al.}~\cite{OEDERLESS_POOLING_2014}
utilize CNN to extract local patch features at multiple scale, and
then pool them to image-level features. Their method harvests deep
activation features and obtains significant performance on many datasets.
Liu \textit{et al.}~\cite{CROSS_CONV_LAYERS} also reveal that the
activations of convolutional layers are useful image representations.

\textcolor{black}{Although directly using activations of a pre-trained
model has good performance for many tasks, fine-tuning a pre-trained
model on one target dataset can further improve the performance by
making the features more specific to the target task~\cite{RICH_FEATURE,DEVIL,ANALYSIS,RGB_D_FT,KTCNN_2008,LFROM_FEW_S,FT_Factors_16,FT-Biomedical}.
Girshick \textit{et al.} \cite{RICH_FEATURE} remove the specific
1000-category classification layer of a CNN which is trained on ImageNet,
and replace it with a randomly initialed (N+1)-category classification
layer (where N is the number of categories in the target data set,
and the extra one category is for the background). To learn the new
parameters, they retrain the modified model on the extracted region
proposal dataset, with unchanged parameters initialized from the pre-trained
one. Their work indicates that fine-tuning is fruitful. Chafield \textit{et al.}
\cite{DEVIL} present rigorously comparisons of the hand-crafted features,
CNNs-based features, and CNNs-based features after fine-tuning on
many datasets. Their work indicates that the fine-tuned CNNs-based
features outperform the others by a large margin. The work of Agrawal
\textit{et al.}\cite{ANALYSIS} and Gupta \textit{et al.}~\cite{RGB_D_FT}
also demonstrates that fine-tuning a per-trained CNN can significantly
improve the performance.}

In order to employ the \emph{inherent knowledge} of a CNN which trained
on a big source dataset, it is intuitive to apply fine-tuning as it
is a reasonable and effective way to make full use of a per-trained
model and alleviate the scarce of training data. In this paper, we
explore factors that influence the performance of fine-tuning, as
illustrated in Fig.~\ref{fig:Factors-that-influence}. There are
many factors that influence the performance of fine-tuning, which
include parameters for the retraining procedure (e.g the initial learning
rate of fine-tuning), the distribution of the source and target data
(e.g., the number of categories in the source dataset, the distance
between the source and target datasets) and so on. To address the
problem of how to fine-tune properly to obtain satisfactory performance
on different classification tasks with various target datasets, in
this work, we quantitatively and qualitatively analyze these factors
respectively. We artificially select different source and target datasets
with different constraints, and then conduct fine-tuning on these
datasets. To the best of our knowledge, this is the first time to
investigate these factors systematically. The main empirical observations
include the following aspects:

\textcolor{black}{First, with roughly the same distance to the target
dataset, the bigger of the number of categories in the source dataset,
the better performance fine-tuning obtains. The pre-trained procedure
works as a regularization to make a better generalization from the
source dataset. For a fixed target dataset, the more categories the
source dataset has, the more knowledge is learned. More categories
make the pre-trained model have better generalization ability, bringing
better performance of fine-tuning.}

\textcolor{black}{Second, when the source dataset is fixed, the performance
of fine-tuning increases with more training examples exposed to the
retraining of the pre-trained model. Whtat is more, the gain of fine-tuning
versus training network from scratch decreases with the increase of
retraining examples. With more training examples, the parameters in
different models both are adjusted to fit the target data, so the
difference decreases.}

\textcolor{black}{At last, we manually select source datasets which
contain the same number of categories but have different similarity
to the target dataset. The results show that the more similar between
the source and target datasets, the better performance fine-tuning
obtains. We analyze the characteristic of different models at both
filter-level and layer-level, and show their sensitivities to the
target dataset. We also qualitatively show the differences among these
different models.}

In the following we give a brief overview of related work in Section\ \ref{sec:Related-Work}.
Section\ \ref{sec:Experimental-Setup} and Section\ \ref{sec:The-factors-that}
present our experimental settings and results. At last, we give our
conclusion in Section\ \ref{sec:Conclusion}.

\section{Related Work\label{sec:Related-Work}}

\textcolor{black}{Our work is related to the work of exploring the
transferability of pre-trained deep models. In this section, we briefly
review the related work about unsupervised pre-training, domain adaptation
and transfer learning.}

\subsection{Unsupervised Pre-training}

Greedy layer-wise pre-training of unsupervised learning followed with
global fine-tuning of supervised learning is an essential component
for deep learning, as the pre-training procedure can introduce useful
prior knowledge. The training strategies for deep models like deep
belief networks (DBN)\ \cite{DBN_2006}, stacked auto-encoders (SAE)\ \cite{SAE_2007}
and stacked denoising auto-encoders (SDAE)\ \cite{SDAE_2008} are
both based on such a similar approach: \textcolor{black}{First, each
layer learns a transformation of its input independently. Second,
all the parameters are fine-tuned with respect to a training criterion.
C}ompared with randomly initialized approaches, the gain obtained
through pre-trained is impressive. The effect of unsupervised pre-training
can be explained as regularization, and it guides the learning towards
basins of attraction of minima that support better generalization
from the training examples~\cite{UNSUPERVISED_PRE_2010}. Whereas,
in the conventional training procedure of CNNs, there is no unsupervised
pre-training stage. However, when fine-tuning a pre-trained CNN model
on a target dataset, we can regard the supervised pre-training as
a substitution of the unsupervised pre-training. In this paper, we
empirically show the influence of various factors when conducting
fine-tuning. 

\subsection{Domain Adaptation}

Domain adaption focus on how to deal with data sampled from different
distributions~\cite{DOMIAN_D_15,DA_2010} (e.g., from product shot
images to real world photos), thus compensating for their mismatch.
Since the theoretical analysis by \cite{Domain_Analysis_2007}, there
have been a lot of research related with this area~\cite{DLID_2013,AD_SENTIMENT_2011,MAXINIZING_FOR_DI}.
Glorot \textit{et al.}~\cite{AD_SENTIMENT_2011} demonstrate that
the features extracted from SDAE are beneficial for the domain adaptation.
Chopra \textit{et al.}~\cite{DLID_2013} propose a method which
learns image representations with respect to domain shift by generating
a lot of intermediate datasets, which are obtained by interpolating
between the source dataset and the target dataset. More similar to
our work, Tzeng \textit{et al.}~\cite{MAXINIZING_FOR_DI} propose
a new CNN with a new adaptation layer and an auxiliary domain confusion
loss to learn domain-invariant representation. Long\ \textit{et al.}~\cite{Domian_2015_long}
propose Deep Adaptation Networks (DANs) to learn transferable features
with statistical guarantees. Their proposed method can also scale
linearly by unbiased estimate of kernel embedding.\textcolor{black}{{}
The aim of our work is to study the factors involved to fine-tuing
but not to obtain the best performance. We focus on one fixed architecture
(i.e., AlexNet) and conduct experiments on subdatasets which are manually
selected from the ImageNet\ \cite{ImageNet-2015-ijcv} dataset. Our
experiments are dedicated to analyzing the factors that influence
the performance of fine-tuning.}

\subsection{Transfer Learning}

Transfer Learning focuses on the possibility to utilize useful knowledge
learned from a source task to master a target task~\cite{TF2010}.
Different from domain adaptation, an essential requirement for successful
knowledge transfer is that the source domain and the target domain
should be closely related\ \cite{DOMIAN_D_15,D_T_Difference_2014}.
\textcolor{black}{Transfer learning has received much attention recently
and many approaches based on CNNs have been proposed in the computer
vision community~\cite{KTCNN_2008,TRANS_ADAPTATON,TRANS_FT,TF_FACTORSs,TL_2018}.
The knowledge transferred from CNNs can be categorized into two kinds.
The first one is in the form of feature representations. These approaches
use a per-trained CNN as a feature extractor and then directly utilize
these extracted features for the target task \cite{DECAF_2014,DEVIL,OFF_SHELF_2014}.
The second one is in the form of model parameters.} Oquab \textit{et al.}
\cite{TRANS_ADAPTATON} propose a framework that remove the output
layer of CNN and add an additional module formed by two fully connected
layers. Their work indicates that this procedure is an effective way
to use the knowledge in a per-trained model. Yosinski \textit{et al.}~\cite{TRANS_FT}
study the transferability of CNNs in each layer. Wang \textit{et al.}~\cite{FT_Brain}
reuse the parameters of a pre-trained model and added new units to
increase its depth and width. Most similar to our work, Azizpour \textit{et al.}~\cite{TF_FACTORSs}
investigate many factors affecting the transferability of CNNs such
as the width and depth of different architectures, the diversity and
density of training data and so on. While fine-tuning is just one
investigated issue in their work. In this work, we systematically
analyze the factors that affect the performance of fine-tuning based
on one model with a fixed structure and shed light on the inner working
mechanism of fine-tuning.

\section{Experimental Setup\label{sec:Experimental-Setup}}

The architecture of the convolutional neural network in our experiments
is almost the same with the one proposed by Krizhevsky \textit{et al.}~\cite{CNNs}.
It has five convolutional layers and three fully-connected layers.
Table \ref{tab:The-architecture-of} details the structure and parameters
of it. In our experiments, we train and fine-tune the CNN with the
Caffe open-source framework~\cite{CAFFE}.

\begin{figure}
\noindent\begin{minipage}[t]{1\columnwidth}%
\begin{center}
\includegraphics[width=0.92\textwidth]{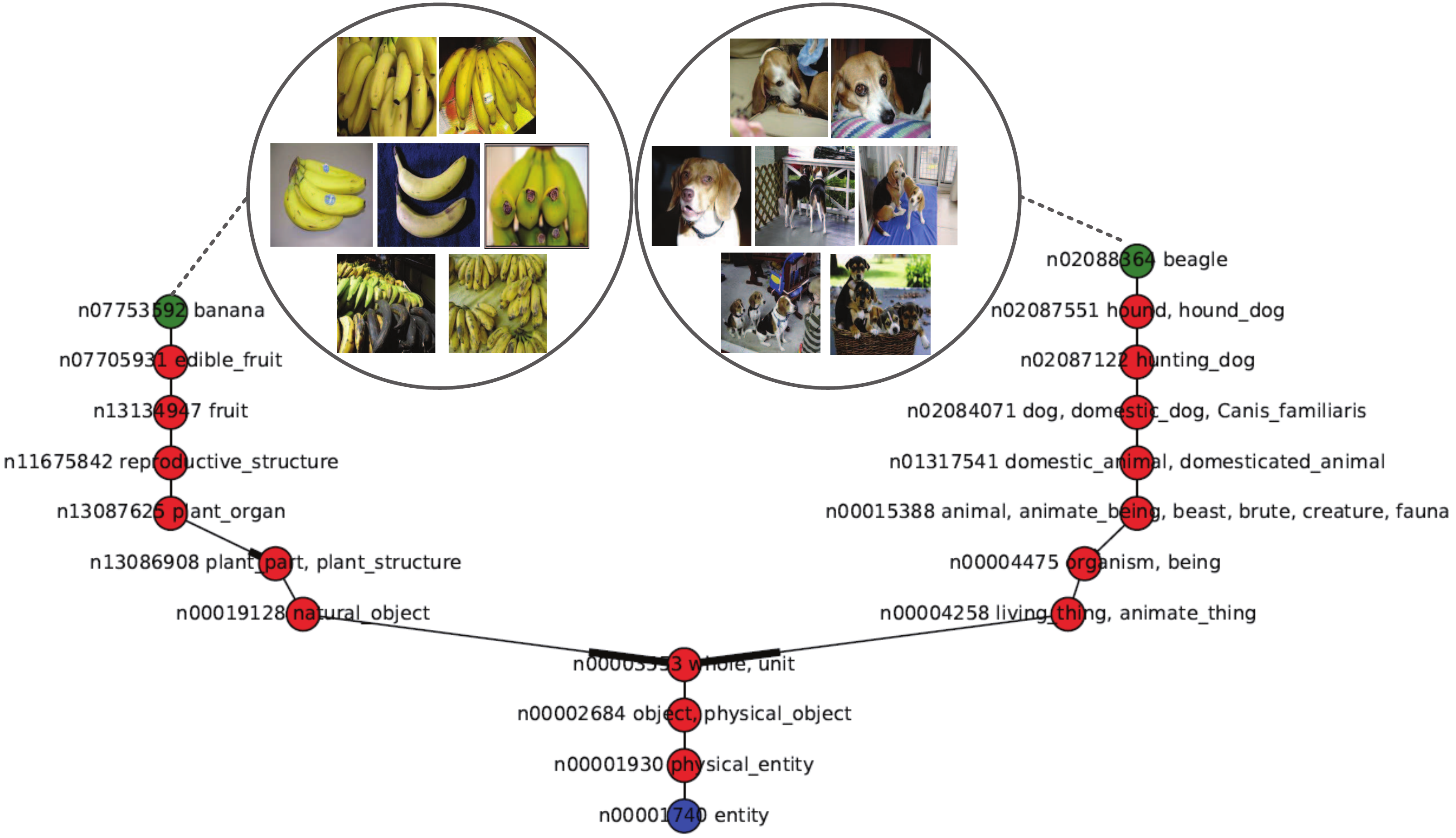}
\par\end{center}%
\end{minipage}

\caption{\textcolor{black}{The parent classes of the banana class and the beagle
class. The green leaf nodes are the target classes, and the blue root
node is the entity class which is the root of WordNet. The red nodes
are the middle classes through which the leaf nodes can reach the
root (Best viewed in color).}\textcolor{red}{\label{fig:Beagle-and-Banana}}}
\end{figure}

\begin{figure*}
\begin{centering}
\includegraphics[width=1\textwidth]{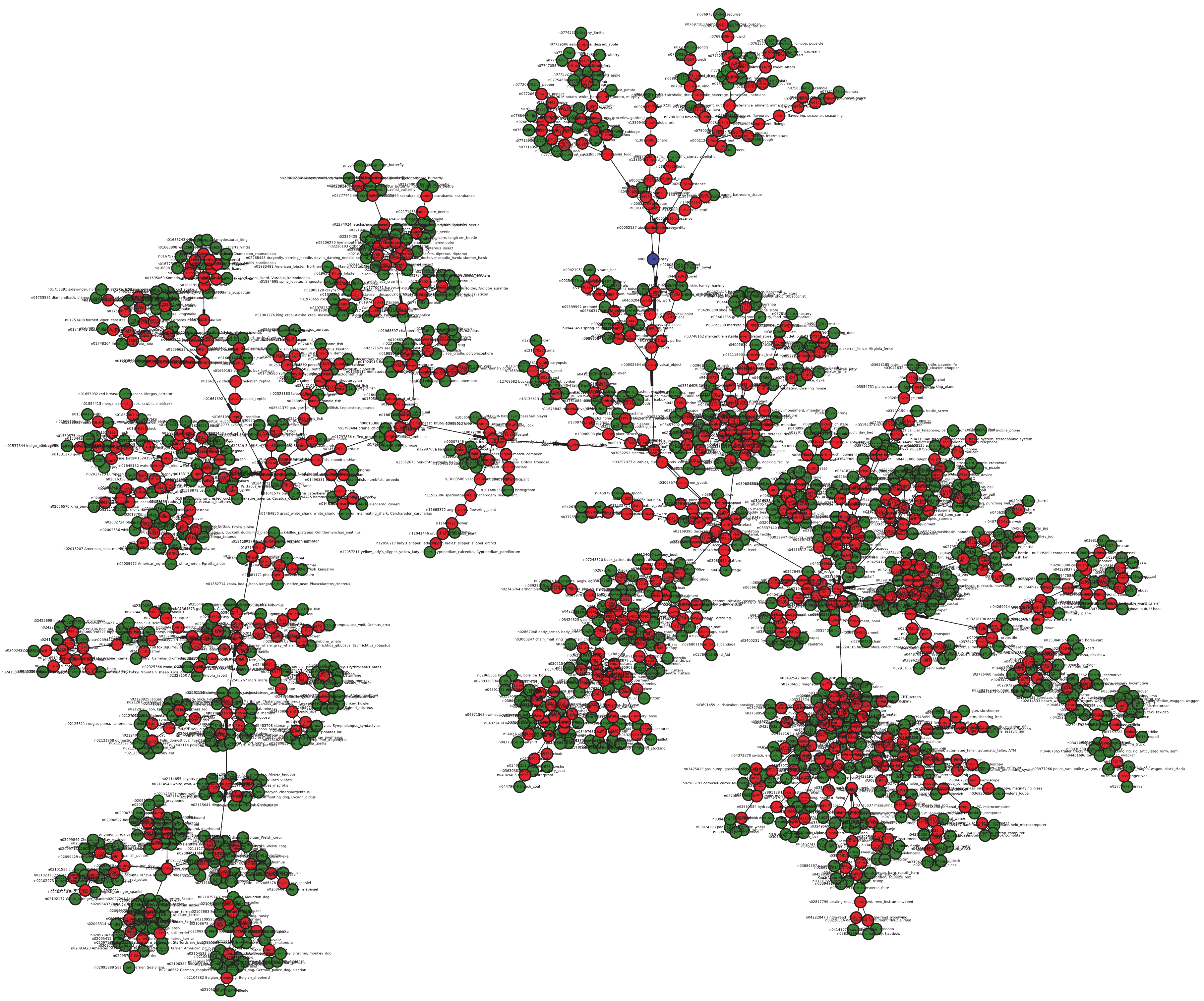}
\par\end{centering}
\centering{}\caption{\textcolor{black}{The tree structure of 1000 ImageNet classes. The
green leaf nodes are the classes in the 1000 ImageNet classes. The
red nodes are the middle classes through which the leaf nodes can
reach the root. The blue root node is the entity class which is the
root of WordNet (Best viewed in color).}\textcolor{red}{\label{fig:The-structure-of}}}
\end{figure*}

\begin{table*}
\begin{centering}
\caption{\textcolor{black}{The architecture of the CNN used in our experiments.
C indicates convolutional layer. R indicates nonlinear layer with
the activation function of Relu. P indicates max pooling layer. N
indicates normalization layer. FC indicates fully-connected layer.}
\label{tab:The-architecture-of}}
\par\end{centering}
\setlength{\tabcolsep}{0.04pt}
\centering{}{\footnotesize{}}%
\begin{tabular}{c|cccccccccccccccccccc}
\hline 
{\footnotesize{}layer} & {\footnotesize{}1} & {\footnotesize{}2} & {\footnotesize{}3} & {\footnotesize{}4} & {\footnotesize{}5} & {\footnotesize{}6} & {\footnotesize{}7} & {\footnotesize{}8} & {\footnotesize{}9} & {\footnotesize{}10} & {\footnotesize{}11} & {\footnotesize{}12} & {\footnotesize{}13} & {\footnotesize{}14} & {\footnotesize{}15} & {\footnotesize{}16} & {\footnotesize{}17} & {\footnotesize{}18} & {\footnotesize{}19} & {\footnotesize{}20}\tabularnewline
\hline 
{\footnotesize{}name} & {\footnotesize{}C1} & {\footnotesize{}R1} & {\footnotesize{}P1} & {\footnotesize{}N1} & {\footnotesize{}C2} & {\footnotesize{}R2} & {\footnotesize{}P2} & {\footnotesize{}N2} & {\footnotesize{}C3} & {\footnotesize{}R3} & {\footnotesize{}C4} & {\footnotesize{}R4} & {\footnotesize{}C5} & {\footnotesize{}R5} & {\footnotesize{}P5} & {\footnotesize{}FC6} & {\footnotesize{}R6} & {\footnotesize{}FC7} & {\footnotesize{}R7} & {\footnotesize{}FC8}\tabularnewline
{\footnotesize{}type} & {\footnotesize{}C} & {\footnotesize{}R} & {\footnotesize{}P} & {\footnotesize{}N} & {\footnotesize{}C} & {\footnotesize{}R} & {\footnotesize{}P} & {\footnotesize{}N} & {\footnotesize{}C} & {\footnotesize{}R} & {\footnotesize{}C} & {\footnotesize{}R} & {\footnotesize{}C} & {\footnotesize{}R} & {\footnotesize{}P} & {\footnotesize{}FC} & {\footnotesize{}R} & {\footnotesize{}FC} & {\footnotesize{}R} & {\footnotesize{}FC}\tabularnewline
{\footnotesize{}channels} & {\footnotesize{}96} & {\footnotesize{}96} & {\footnotesize{}96} & {\footnotesize{}96} & {\footnotesize{}256} & {\footnotesize{}256} & {\footnotesize{}256} & {\footnotesize{}256} & {\footnotesize{}384} & {\footnotesize{}384} & {\footnotesize{}384} & {\footnotesize{}384} & {\footnotesize{}256} & {\footnotesize{}256} & {\footnotesize{}256} & {\footnotesize{}4096} & {\footnotesize{}4096} & {\footnotesize{}4096} & {\footnotesize{}4096} & {\footnotesize{}1000}\tabularnewline
{\footnotesize{}size} & {\footnotesize{}11} & {\footnotesize{}-} & {\footnotesize{}3} & {\footnotesize{}-} & {\footnotesize{}5} & {\footnotesize{}-} & {\footnotesize{}3} & {\footnotesize{}-} & {\footnotesize{}3} & {\footnotesize{}-} & {\footnotesize{}3} & {\footnotesize{}-} & {\footnotesize{}3} & {\footnotesize{}-} & {\footnotesize{}3} & {\footnotesize{}-} & {\footnotesize{}-} & {\footnotesize{}-} & {\footnotesize{}-} & {\footnotesize{}-}\tabularnewline
{\footnotesize{}parameters} & {\footnotesize{}11712} & {\footnotesize{}-} & {\footnotesize{}-} & {\footnotesize{}-} & {\footnotesize{}6656} & {\footnotesize{}-} & {\footnotesize{}-} & {\footnotesize{}-} & {\footnotesize{}3840} & {\footnotesize{}-} & {\footnotesize{}3840} & {\footnotesize{}-} & {\footnotesize{}2560} & {\footnotesize{}-} & {\footnotesize{}-} & {\footnotesize{}37752832} & {\footnotesize{}-} & {\footnotesize{}16781312} & {\footnotesize{}-} & {\footnotesize{}4097000}\tabularnewline
\hline 
{\footnotesize{}percentage} & {\footnotesize{}0.01997} &  &  &  & {\footnotesize{}0.01135} &  &  &  & {\footnotesize{}0.006546} &  & {\footnotesize{}0.006546} &  & {\footnotesize{}0.004364} &  &  & {\footnotesize{}64.36} &  & {\footnotesize{}26.61} &  & {\footnotesize{}6.984}\tabularnewline
\hline 
\end{tabular}{\footnotesize \par}
\end{table*}

For training the model on the source dataset, the learning choices
are the same with \cite{CNNs}. \textcolor{black}{The weights in each
layer are initialized from a zero-mean gaussian distribution with
deviation 0.01, except the deviations of layer fc6 and fc7 are 0.005.}
The neuron biases in conv2, conv4 and conv5, as well as all the fully-connected
layers are initialized with the constant 1, while the biases in other
layers are initialized with the constant 0. The momentum is 0.9 and
the weight decay is 0.0005. The learning rate at the beginning is
0.01 (for training models from scratch) and after every 20 cycles
it is divided by the constant 10. We train the CNN for roughly 100
cycles. To reduce overfitting, first there are two dropout layers
with a dropout ratio of 0.5 followed the layer of fc6 and fc7. Second,
we randomly crop $227\times227$ pixels from the $256\times256$ input
image and randomly mirror it in each forward and backward processing. 

For fine-tuning the pre-trained model on the target dataset, we remove
the last fully-connected layer which is specific to the source task
and replace it with a new randomly initialized one with $C$ units
(where $C$ is the number of categories in the target task). After
this we use stochastic gradient descent (SGD) to continuously train
(i.e., retrain) the modified model on the target dataset. In contrast
to the work in\ \cite{TRANS_FT}, the parameters copied from the
pre-trained model are optimized with respect to the target task in
our experiments, while these transferred parameters are fixed during
retraining in their work.

We analyze the factors that affect fine-tuning mainly on the 1000
ImageNet classes~\cite{ImageNet-2015-ijcv} dataset (ILSVRC2012),
as shown in Fig.\ \ref{fig:The-structure-of}. It totally has $1.28$
million images with abundant semantic diversity and density. In order
to find subdatasets with different distances (for example, A is a
dataset which just contains domestic dogs, B is a dataset which contains
many other animals such as cats, sheep, while C is a dataset which
contains many traffic instruments. We define the distance between
A and B is smaller than the distance between A and C qualitatively,
as objects in A and B both have eyes, legs and tails while objects
in C do not have), we draw out the tree structure of 1000 ImageNet
classes according to WordNet~\cite{WORDNET}. For every category
in the 1000 classes, it belongs to one synset in WordNet. We can find
its parent classes recursively until reaching the node of entity which
is the root of the WordNet. For example, the parent classes of the
``n02088364 beagle'' class and the class of ``n07753592 banana''
are showed in Fig.\ \ref{fig:Beagle-and-Banana}. The tree structure
of the whole 1000 classes is illustrated in Fig.\ \ref{fig:The-structure-of}.
In the following of our paper, we select subdatasets from the 1000
ImageNet classes with different constraints.

\section{The factors that affect the performance of Fine-tuning\label{sec:The-factors-that}}

\textcolor{black}{There are many factors that affect the performance
of fine-tuning, such as the number of categories in the source dataset,
the distance between the source and target dataset, the number of
categories in the target dataset, the number of training examples
per category in the target dataset and so on.}\textcolor{red}{{} }In
our paper, we individually analyze the effect of each factor. For
training models from scratch and fine-tuning models from pre-trained
ones, we use all the images (which belong to the selected categories)
in the training set of ImageNet. We first train a set of models on
different datasets and then fine-tune these model on the target datsets.
\textcolor{black}{For testing, we use 50 images per category, which
are the same with the validation set of ImageNet.}

\subsection{The Initial Learning Rate of Fine-tuning}

\begin{figure*}
\noindent\begin{minipage}[t]{1\columnwidth}%
\begin{center}
\includegraphics[width=0.84\textwidth]{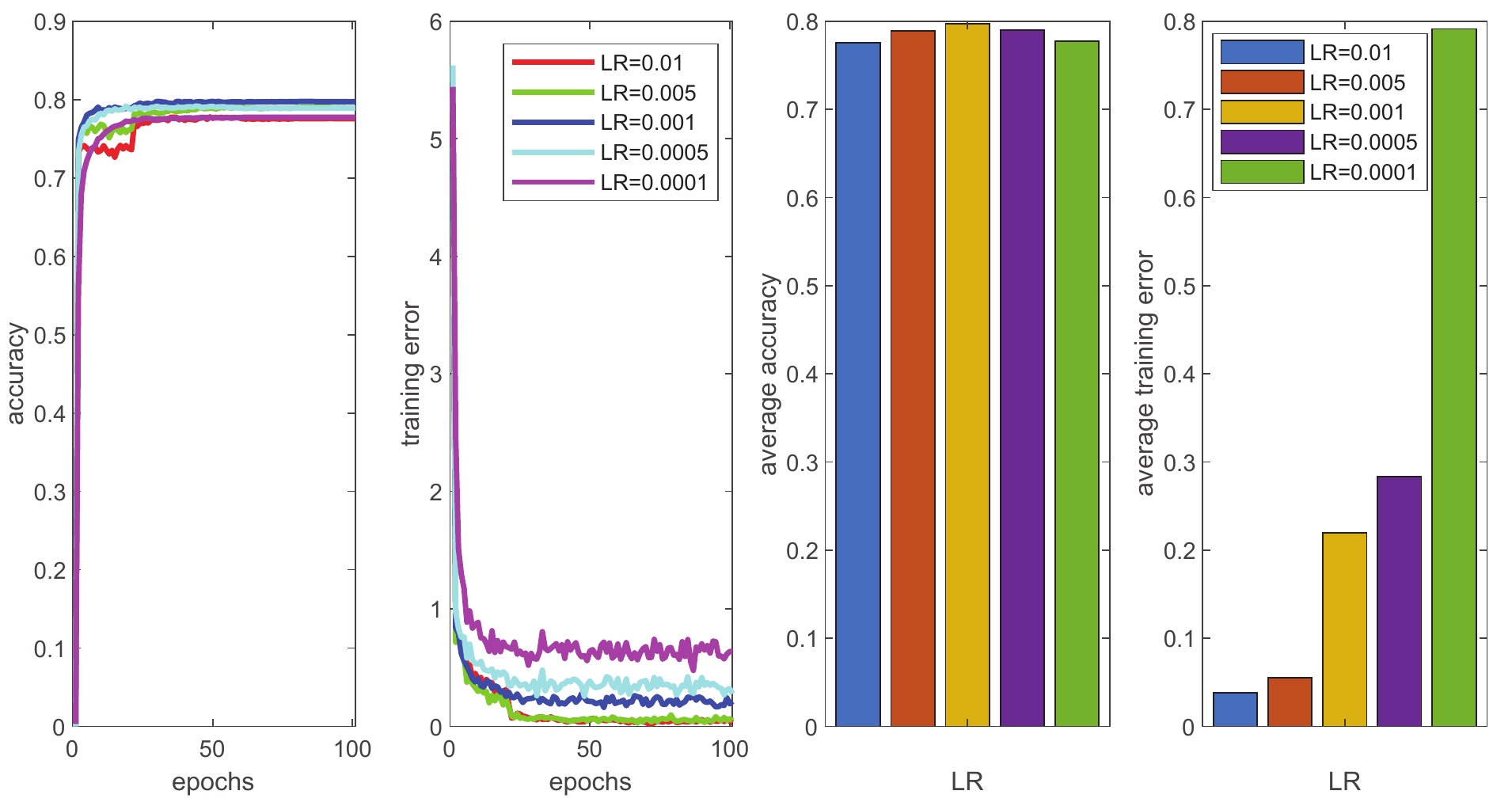}
\par\end{center}%
\end{minipage}

\caption{\textcolor{black}{The effect of the initial learning rate on the performance
of fine-tuning. The first column is the performance on the test set
when fine-tuning the pre-trained model with different initial Learning
Rates (LRs). The second column is the training error (i.e., training
loss). The third and fourth columns are the average accuracy and the
average training error. The experiments are conducted independently
for five times.}\label{fig:The-effect-o-fLR}}
\end{figure*}

Fine-tuning is implemented by initializing a network with parameters
optimized on one source dataset except the last task-specific layer
which is randomly initialized. Then, using the target training samples,
the parameters of the network are updated. When using stochastic gradient
descent to optimized the weights, the initial Learning Rate (LR) has
big influence on the learning procedure. As previous work has mentioned~\cite{RICH_FEATURE,TF_FACTORSs},
the initial LR at the beginning should be smaller than the initial
LR used to optimize the wights on the source dataset. This strategy
ensures the prior knowledge learned in the source dataset are not
clobbered. 

\textcolor{black}{However, to what extent does the initial LR affect
the fine-tuning performance? In this section, we experimentally investigate
this factor. We randomly select 100 classes from the 1000 ImageNet
classes as the target dataset. And then, we train the model from scratch
on the remaining 900 categories. After this, we fine-tune the pre-trained
model on 100-class target dataset with decreasing initial Learning
Rates: 0.01, 0.005, 0.001, 0.0005 and 0.0001. The experiments are
conducted independently for five times (i.e., five target datasets
and five source datasets). The results are shown in Fig.~\ref{fig:The-effect-o-fLR}.
When the initial LR is set to 0.02 or bigger, the learning procedure
does not converge as we have tried in our experiments. So we set 0.01
as the biggest initial LR. The remarkable observations are: (1) Bigger
initial LRs bring smaller training errors. (2) Starting with a big
initial LR, there are some bumps in the early training epochs. This
indicates the basins of attraction of minima are reconstructed. Though
the training error is much smaller, the generalization ability is
worse, as the test accuracy is not good enough compared with a moderate
initial LR. (3) When starting with a very small LR, the training error
is much bigger and the test accuracy is smaller. This indicates the
learning procedure is not plenitudinous. When starting with the initial
LR of 0.001, the learning procedure is mild, and it obtains the best
performance. So in the rest of our experiments, we fixed the initial
LR with 0.001.}

\subsection{The Number of Categories in the Source Dataset\label{subsec:The-number-of_categotris}}

\textcolor{black}{Previous work has suggested that increasing the
training data can significantly improve the performance~\cite{TRANS_ADAPTATON,TF_FACTORSs}.
In this section, we investigate the effect of the number of categories
in the source dataset for fine-tuning. First we randomly select 100
categories from ImageNet as the fixed target dataset. And then, we
randomly select 800, 500, 200, 100, and 80 categories respectively
from the remaining categories as different source datasets. Thus,
the distances between these source datasets and the target dataset
are roughly the same.}

\begin{figure*}
\noindent\begin{minipage}[t]{1\columnwidth}%
\begin{center}
\includegraphics[width=0.88\textwidth]{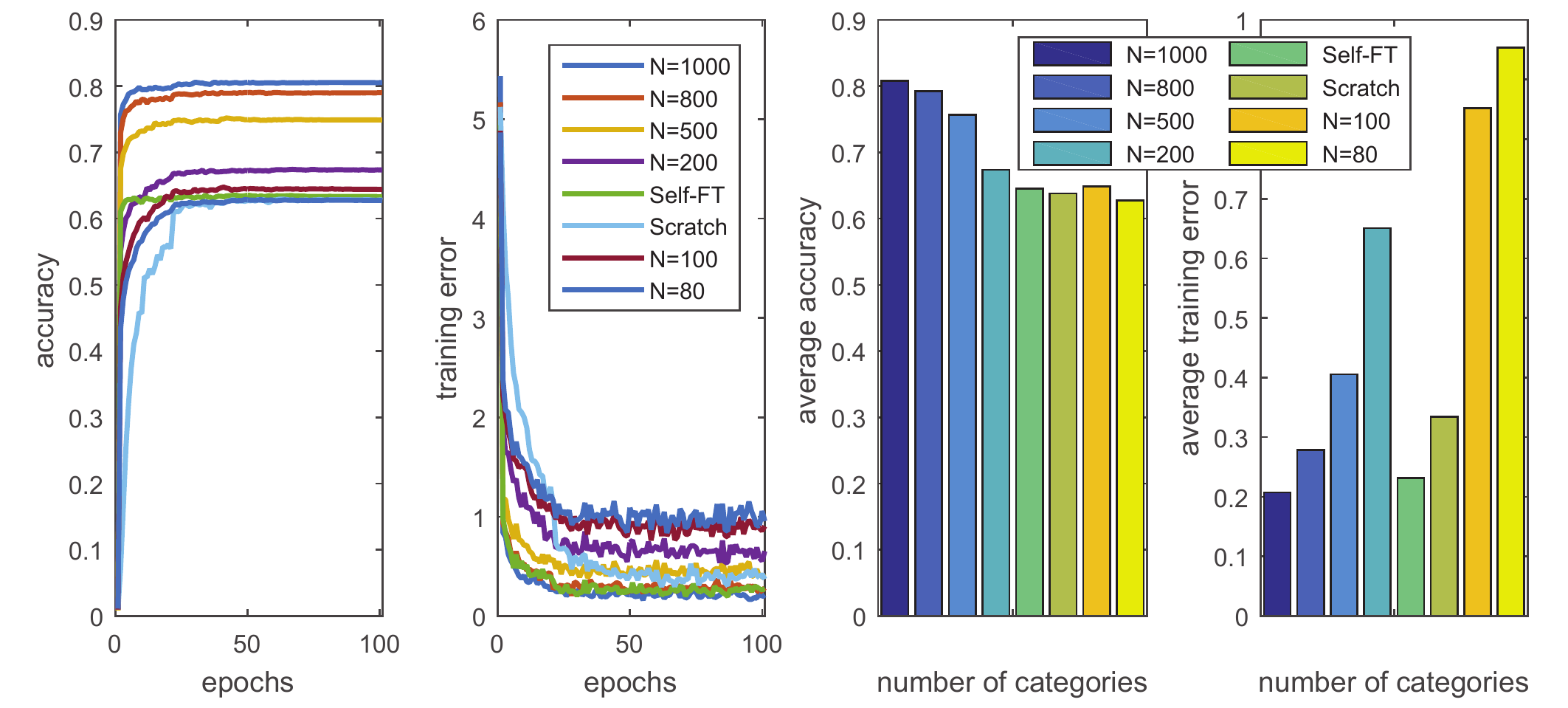}
\par\end{center}%
\end{minipage}

\caption{\textcolor{black}{The effect of the number of categories in the source
dataset when fine-tuning from different pre-trained models on the
fixed target dataset.\label{fig:The-effect-of_categoreis} The first
column is the accuracy and the second column is the training error
(i.e., training loss). The third and fourth columns are the average
values. We run the experiments with the same settings for five times.}}
\end{figure*}

After obtaining different kinds of pre-trained models which are trained
on these selected source datasets, we fine-tune these models on the
fixed target dataset.\textcolor{black}{{} Fig.~\ref{fig:The-effect-of_categoreis}
is a plot of the accuracy and training error when fine-tuning different
pre-trained models on a fixed target dataset. We also conduct the
experiments independently for five times and Table~\ref{tab:The-performance-different}
is the the average performance. There are several observations can
be drawn from these results. }

First, when we fine-tune a pre-trained model on a dataset, it dramatically
converges faster than training it from scratch, especially when the
sour dataest has many categories. The explanation is that the model
pre-trained on dataset which contains enough categories has a better
basin of attraction of minima, and its parameters can be efficiently
adapted to new tasks with little changes. 

Second, as expected, a network trained on more categories, it learned
more knowledge, and the generalization ability of the network is much
better. As we can seen in Fig.~\ref{fig:The-effect-of_categoreis},
with the decrease of the number of categories in the pre-trained model,
the performance of fine-tuning decreases and the training error goes
larger. For example, the model trained on 1000 categories (N=1000)
has the best performance and the training error is the smallest. When
the pre-trained number of categories is 500 (N=500), the training
error is larger and the performance decreases. But it is obvious that
even though the training error (N=100) is larger, the performance
is still better than the performance of training the network from
scratch. When fine-tuning from itself (continued training on the target
dataset itself), the training error goes smaller, but the test accuracy
is almost the same with the accuracy of training the network from
scratch.

At last, when fine-tuning from a deficient model which trained on
a source dataset which only has s small number of categories (N=80),
the performance is even worse than training the model from scratch.
This can be attributed to the perishing generalization ability of
the pre-trained model and the small learning rate of fine-tuning. 

\textcolor{black}{With roughly the same distance to the target dataset,
the bigger of the number of categories in the source dataset, the
better performance fine-tuning obtains, as shown in the results. It
can be explained as that the procedure of pre-trained a model works
as a regularization, making better generalization from the source
dataset. The more knowledge learned in the source dataset, the better
generalization ability the pre-trained model has.}

\begin{table}
\caption{The performance (mean and variance) of fine-tuning different pre-trained
models on the fixed target dataset.\label{tab:The-performance-different}
We conduct the experiments respectively for five times.}

\setlength{\tabcolsep}{2.6pt}
\centering{}%
\begin{tabular}{c|cccc}
\hline 
\multicolumn{1}{c|}{Number} & 1000 & 800 & 500 & 200\tabularnewline
\hline 
Accuracy(\%) & 80.94$\pm$0.75 & 79.25$\pm$0.23 & 75.97$\pm$1.17 & 66.76$\pm$1.71\tabularnewline
\hline 
\hline 
Number & SF-FT & Scratch & 100 & 80\tabularnewline
\hline 
Accuracy(\%) & 63.53$\pm$2.33 & 62.47$\pm$2.50 & 63.67$\pm$1.67 & 61.35$\pm$2.02\tabularnewline
\hline 
\end{tabular}
\end{table}

\subsection{The Distance between the Source and Target Datasets\label{subsec:The-distance-between}}

\begin{figure}
\noindent\begin{minipage}[t]{1\columnwidth}%
\begin{center}
\includegraphics[width=0.54\textwidth]{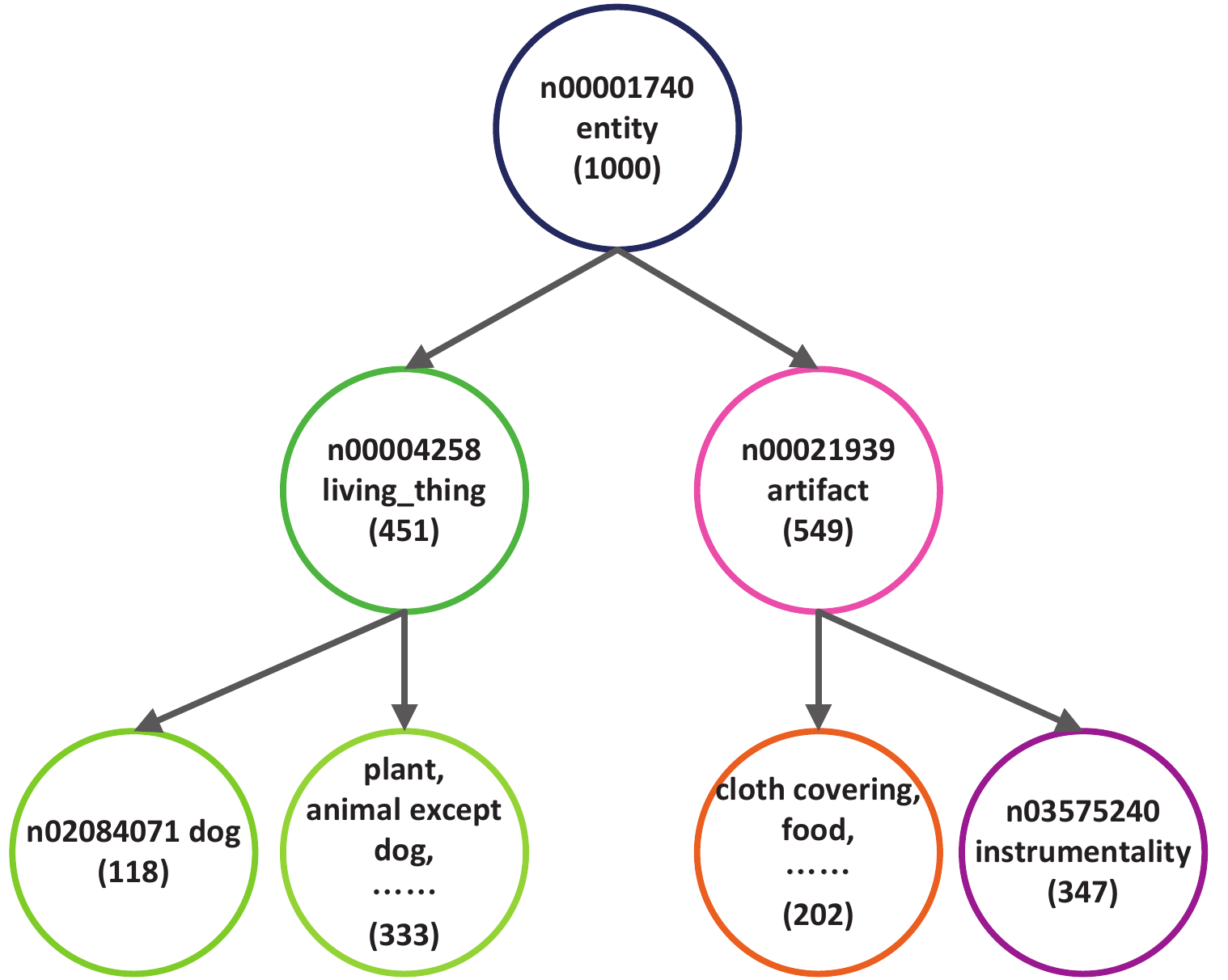}
\par\end{center}%
\end{minipage}

\caption{\textcolor{black}{Different branches of the 1000 categories in ImageNet.\label{fig:Different-branches-of}
The integer in the parenthesis indicates the number of the categories
that the branch contains. The number in the beginning of the node
is the corresponding WordNet ID.} }
\end{figure}

\begin{figure*}
\noindent\begin{minipage}[t]{1\columnwidth}%
\begin{center}
\includegraphics[width=0.92\textwidth]{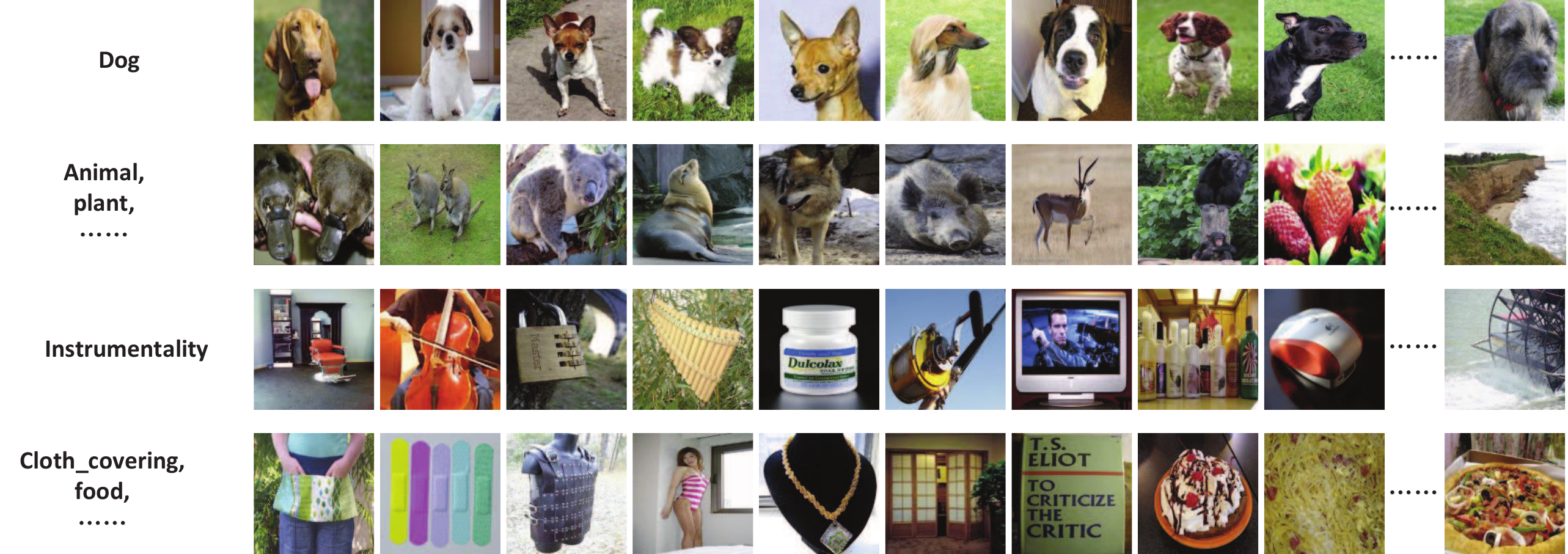}
\par\end{center}%
\end{minipage}

\caption{Images belong to different branches.\label{fig:Branches-of-images}
The first row shows images belong to the super class of dog (e.g.,
rhodesian ridgeback, papillon, and saint bernard). The second row
shows images which are the rest of the living\_thing super class (e.g.,
plant, platypus, kangroo, and wolf) except the dog class. The third
row shows the super class of instrumentality (e.g., lock, fishing
pole, and mouse) and the last row shows images which belong to the
rest of the artifact super class (e.g., cloth\_covering, food) except
the instrumentality class.}
\end{figure*}

\textcolor{black}{It is well known that dataset bias is a very common
issue in visual recognition~\cite{BAIS,MAXINIZING_FOR_DI,TRANS_FT,TF_FACTORSs}.
However, when fine-tuning pre-trained models on a target dataset,
what is the relationship of the corresponding performance and the
distance between the source and target datasets? It is meritorious
to expose what happens when fine-tuning from different pre-trained
models. In this section, we investigate this factor qualitatively
and quantitatively and reveal insights into what happens when conducting
fine-tuning.}

\begin{table}
\caption{The performance of fine-tuning different pre-trained models on the
target dataset (\%).\label{tab:The-performance-of} CNN indicates
the accuracy which is the direct output of the CNN. SVM indicates
the result of the features of layer fc7 and the combination of a SVM
classifier. The performance before fine-tuning is reported in parentheses.
Scratch represents the accuracy of training the model from scratch.}

\centering{}%
\begin{tabular}{cc|ccc}
\hline 
 &  & near\_DOG & far\_DOG & scratch\tabularnewline
\cline{2-5} 
\multirow{2}{*}{DOG} & CNN & 59.56 & 55.94 & 55.62\tabularnewline
 & SVM & 58.26 (44.98) & 55.14 (28.88) & 55.38\tabularnewline
\hline 
\multirow{3}{*}{INSTRU} &  & near\_INSTRU & far\_INSTRU & scratch\tabularnewline
\cline{2-5} 
 & CNN & 67.12 & 61.58 & 54.22\tabularnewline
 & SVM & 66.70 (60.06) & 61.00 (47.54) & 54.00\tabularnewline
\hline 
\end{tabular}
\end{table}

\textcolor{black}{Measuring the distance between two datasets is very
crucial in image recognition. We consider the way used by Yosinski
\cite{TRANS_FT}. More precisely, we manually select nodes in different
branches in the tree structure of the 1000 categories in ImageNet,
thus forming subdatasets which have different distance to a fixed
target dataset, as illustrated in Fig.\ \ref{fig:Different-branches-of}.
For example, the 1000 categories are divided into two branches, one
is living thing which contains 451 categories, and another one is
artifact which contains 549 categories. We first select two fixed
datasets with 100 categories, and then select two datasets with different
distance with 500 categories for each of the target dataset. The procedures
to obtain the datasets are: (1) The dog branch has 118 categories
(e.g., rhodesian ridgeback, papillon, and saint bernard), as shown
in Fig.~\ref{fig:Branches-of-images}. Therfore, 100 categories are
randomly selected from this branch as the first target dataset which
is denoted as DOG. And then, we randomly select 500 categories from
the 549 categories super class of artifact as a source dataset, which
is denoted as far\_DOG. Near\_DOG is composed of the rest of the 400
(1000-100-500) categories and 100 categories randomly selected from
far\_DOG. The distance between near\_DOG and DOG is smaller than the
distance between far\_DOG and DOG, as there are many categories in
near\_DOG which have eyes, legs, tail and fur (e.g., platypus, kangroo,
and wolf) while categories in far\_DOG do not have. (2) The second
target dataset is selected from a more widely distributed way. As
there are 347 categories in the instrumentality super class (e.g.,
lock, fishing pole, and mouse), 100 categories are randomly chosen
as the target dataset which is denoted as INSTRU. The 451 categories
super class of living thing and a random choose of 49 categories from
the super class of cloth covering and food which contain 202 categories
compose far\_INSTRU. Near\_INSTRU contains the rest 400 (1000-100-500)
categories and 100 categories randomly selected from far\_INSTRU. }

The performance of fine-tuning different pre-trained models on the
target dataset is illustrated in Table~\ref{tab:The-performance-of}.
In order to measure the representative and discrimination of different
pre-trained models, we also utilize the features of layer fc7 and
a simple SVM classifier to obtain the classification accuracy on the
target dataset.\textcolor{black}{{} It is can be obtained that the distance
between the source and target datasets has much influence on the performance
of fine-tuning. F}or DOG recognition, the features directly extracted
from the model pre-trained on near\_DOG (44.98\%) is much better than
that of far\_DOG (28.88\%). For the model which is pre-trained on
the source dataset which is close to the target one, it learns prior
knowledge which is useful to classify the target images. So near\_DOG
performs better than far\_DOG. Even more, the performance of fine-tuning
the model which is pre-trained on near\_DOG (58.26) is also much better
than the fine-tuned far\_DOG (55.14), as expected. And the trend is
the same in INSTRU. It is demonstrated that fine-tuning from a near
model yields better performance than a far one. \textcolor{black}{The
results show that the more similar between the source and target datasets,
the better performance fine-tuning obtains.}

\begin{figure*}
\setlength{\tabcolsep}{0.5pt}
\begin{centering}
\begin{tabular}{c}
\includegraphics[width=0.94\textwidth]{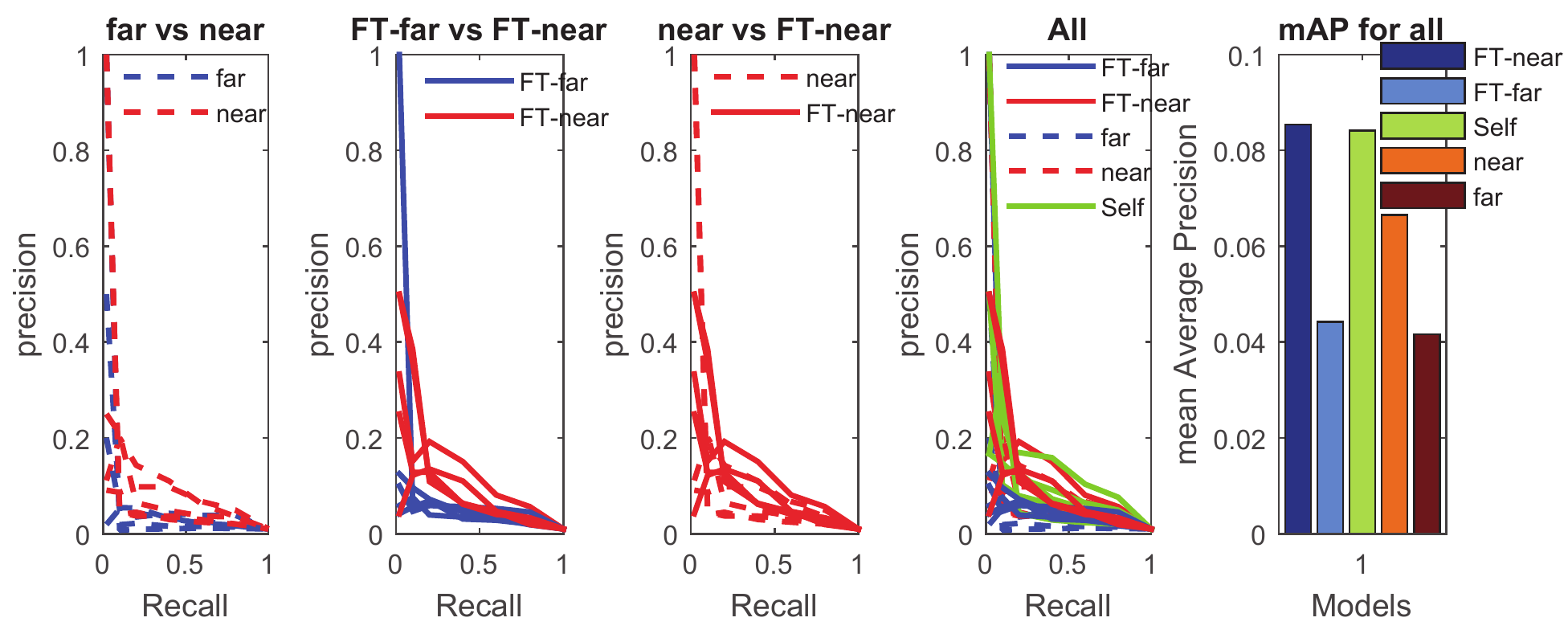}\tabularnewline
\includegraphics[width=0.94\textwidth]{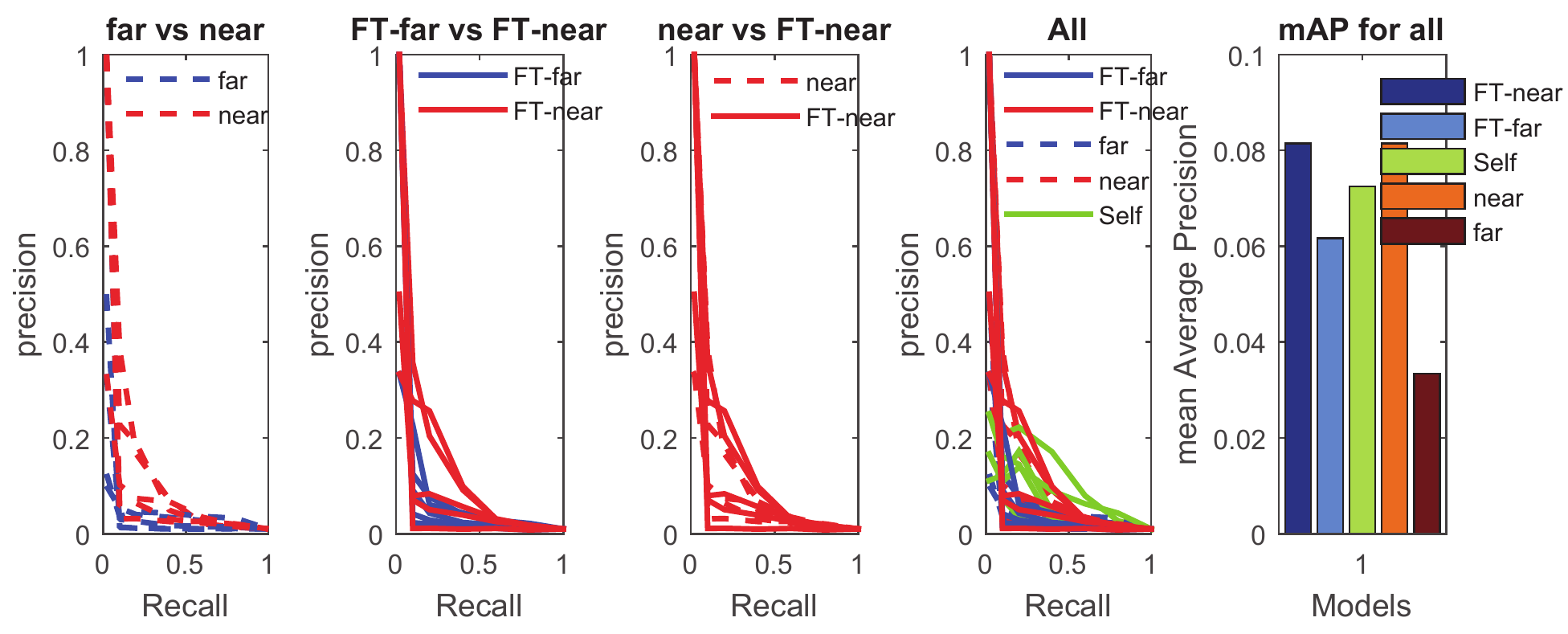}\tabularnewline
\includegraphics[width=0.94\textwidth]{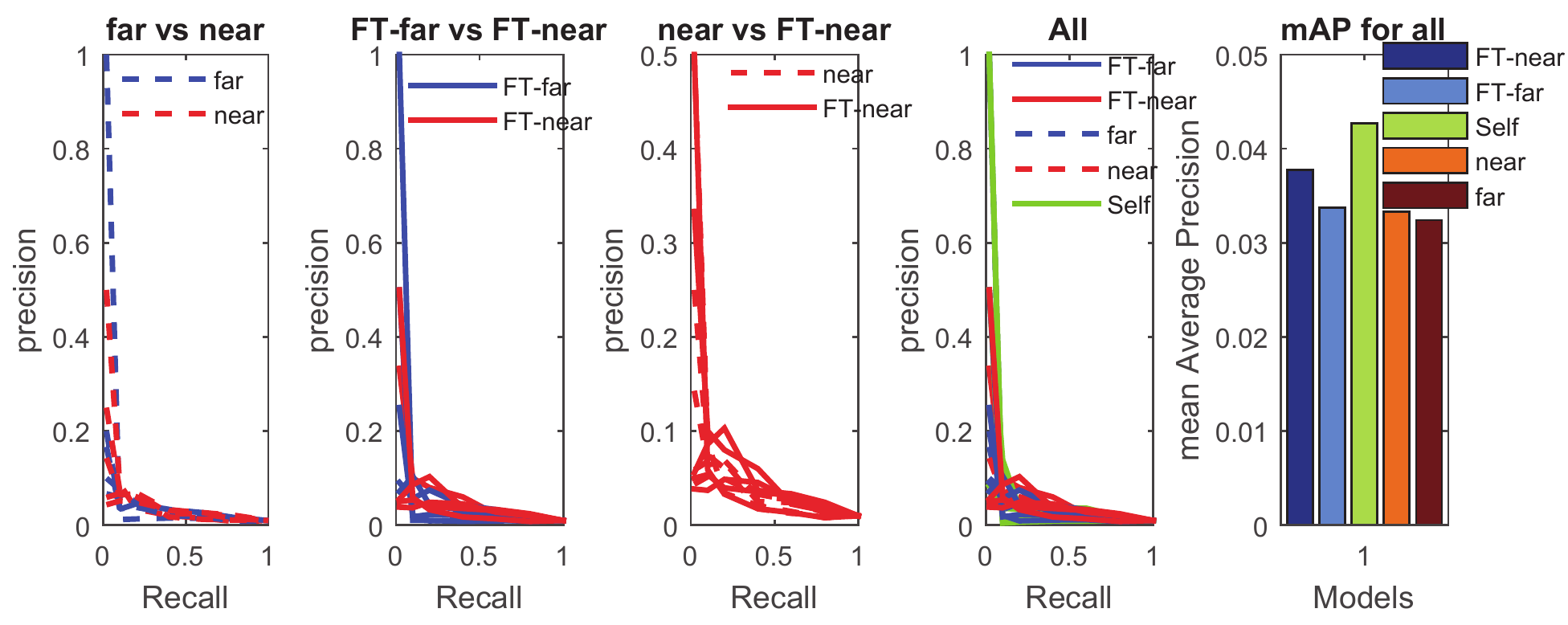}\tabularnewline
\end{tabular}
\par\end{centering}
\caption{\textcolor{black}{The sensitivity of filters in the conv5 layer.}\label{fig:The-sensitivity-of-conv5-1}
The first row is for the class of \emph{garbage truck. }The second
and third rows are for the class of \emph{forklift} and \emph{pencil
box} respectively. The red dash lines in the first column shows the
most top-five filters in the model pre-trained on near\_INSTRU (near)
and the blue dash lines are for far\_INSTRU (far). The red solid lins
in the second cloumn are the most top-five filters after fine-tuning
near on INSTRU (FT-near) and blue solid liens are for (FT-far). Red
solid lines in the third cloumn are for FT-near and red dash lines
are for near. The fourth cloumn shows top-five filters in all the
five models. Specifically, the green solid lines are for the model
trained on INSTRU itself (Self). The last cloumn shows the mean AP
of the five filters for each model.}
\end{figure*}

\begin{figure}
\noindent\begin{minipage}[t]{1\columnwidth}%
\begin{center}
\includegraphics[width=0.72\textwidth]{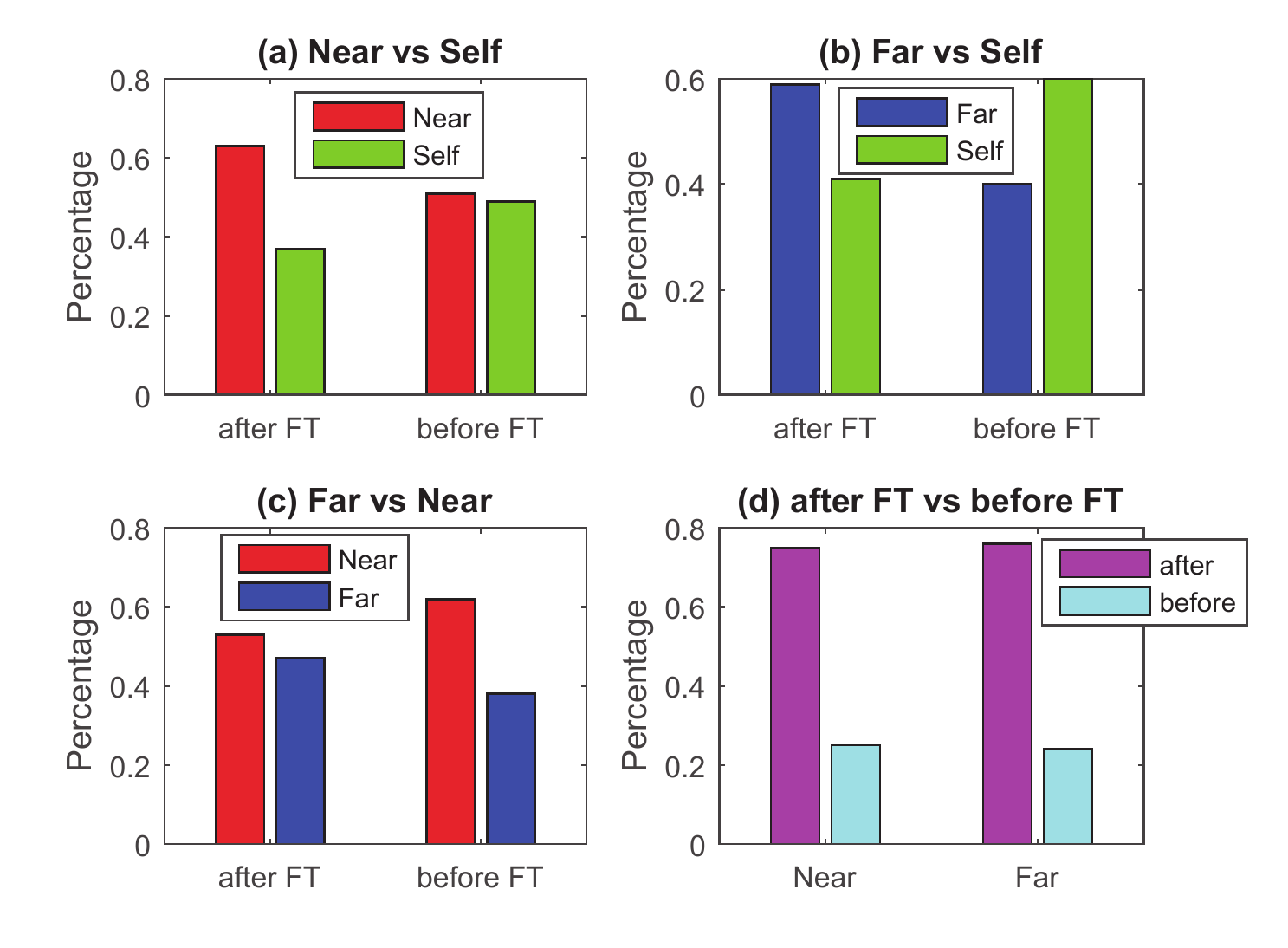}
\par\end{center}%
\end{minipage}

\caption{The statistics of sensitivity percentages for different models on
the 100 categories in INSTRU.\label{fig:The-statistics-of-all} Near
represents the model first trained on near\_INSTRU, and far indicates
the fine model first trained on far\_INSTRU. Self is the model directly
trained on INSTRU from scratch.}
\end{figure}

\subsubsection{The discrimination of individual filters}

The discrimination of different layers indicate the class selectivity
of the group of filters in the particular layer. \textcolor{black}{Even
though most of the features in convolution neural networks are distributed
code, it is also important to measure the characteristic of individual
filters, as cells in human brain have big response to some specific
and complex stimuli, which has close relationships to object recognition
\cite{1972-psy}. }The discrimination of one individual filter on
one class can be regarded as its sensitivity to the class. More precisely,
if a filter is sensitive to one class, the filter should fire strongly
on all images in this class, and at the same time it should also have
small response to images in other classes. 

\textcolor{black}{In order to measure the sensitivity of one filter
for one class, we use the precision and recall curve to compute the
criterion quantitatively, just as Agrawal do in \cite{ANALYSIS}.
Specifically, we analyze the filters on the conv5 layer, as features
of this layer are mid-level image representations, which are the combination
of low layer features and also have less semantic information compared
to the features in fully-connected layers. The features of conv5 are
consisted of responses of 256 filters, which have a size of $256\times6\times6$.
We use max-pooling to transform the activations of spatial grid of
size $6\times6$ to one value, as this strategy just causes a small
drop in recognition performance while brings a shorter and more compact
features. With this implementation, the responses of size $256\times6\times6$
are reduced to $256\times1$, each element of which indicates the
response of one filter. Now, for a set of images, we will get a set
of associated scores. We treat every filter as a classifier and compute
the precision recall curve for each of them.}

\textcolor{black}{For each class, we now can get 256 curves with each
curve correspond to one filter. The curves are computed on the test
set of INSTRU which contains 100 categories, with 50 images per class.}
We sort the filters in descending order of their average precision
values (APs). Fig.~\ref{fig:The-sensitivity-of-conv5-1} shows the
top-fives filters of their precision recall curves for the class of
\emph{garbage trunk, forklift, }and \emph{pencil box} respectively.
For the \emph{garbage truck} class, the top-five filters in the near
model are more sensitive than the filters in model of far, as the
precision recall curves of far are almost under the precision recall
curves of near. In order to compare one individual filter, we use
the mean average precision (mAP) of the selected five filters as the
sensitivity, as shown in last column of Fig.\ \ref{fig:The-sensitivity-of-conv5-1}.\textcolor{red}{{}
}\textcolor{black}{The mAP of near is bigger than that of far. After
fine-tuning, both of the mAPs for far and near obtain increase.}

In order to compare these filters in different models in a global
level, we calculate the number of categories that the filters in one
model has bigger mAP than filters in the others. We use sensitivity
percentages to evaluate the global sensitivity of two models when
comparing them. The bigger of percentage that one model has, the more
sensitive the model is. Fig. \ref{fig:The-statistics-of-all} shows
the sensitivity percentages of different models on the 100 categories
in INSTRU. \textcolor{black}{There are many conclusions can be obtained
from the results.}

\textcolor{black}{First, filters in near are more sensitive to filters
in far. As abundant visual information is put into the near model,
filters in near are more sensitive than filters in self. While far\_INSTRU
has big distance to INSTRU, so the filters in the model trained on
it (far) are more insensitive than the filters in self, filters in
far only have a bigger mAP on 40\% categories, as shown in Fig.~\ref{fig:The-statistics-of-all}
(a) and (b).}

\textcolor{black}{Second, fine-tuning make the filters more sensitive
to the target dataset. For 51\% of the categories in INSTRU, the filters
in the model trained on near\_INSTRU (near) have a bigger mAP compared
with filters in the model trained on INSTRU (self). While after fine-tuning,
the filters in near have a bigger mAP than filters in self on 61\%
of the categories, as shown in Fig.~\ref{fig:The-statistics-of-all}
(a). Meanwhile, Fig.~\ref{fig:The-statistics-of-all} (d) shows that
filters after fine-tuning have bigger mAP compared with filters before
fine-tuning on most of the categories on both near (75\%) and far
(76\%).}

\textcolor{black}{At last, the sensitivity difference between near
and far is big, as they are optimized on different dataset focusing
on different visual patterns. After fine-tuning, they both fire on
the target images, so the difference between them decreases. As shown
in Fig.~\ref{fig:The-statistics-of-all} (c), after fine-tuning,
filter in near have big mAP only on 53\% categories.}\textcolor{red}{{}
}

\begin{figure*}
\noindent\begin{minipage}[t]{1\columnwidth}%
\setlength{\tabcolsep}{0.5pt}
\begin{center}
\begin{tabular}{cc}
\multicolumn{2}{c}{(a) Performance of different layers}\tabularnewline
\multicolumn{2}{c}{\includegraphics[width=0.86\textwidth]{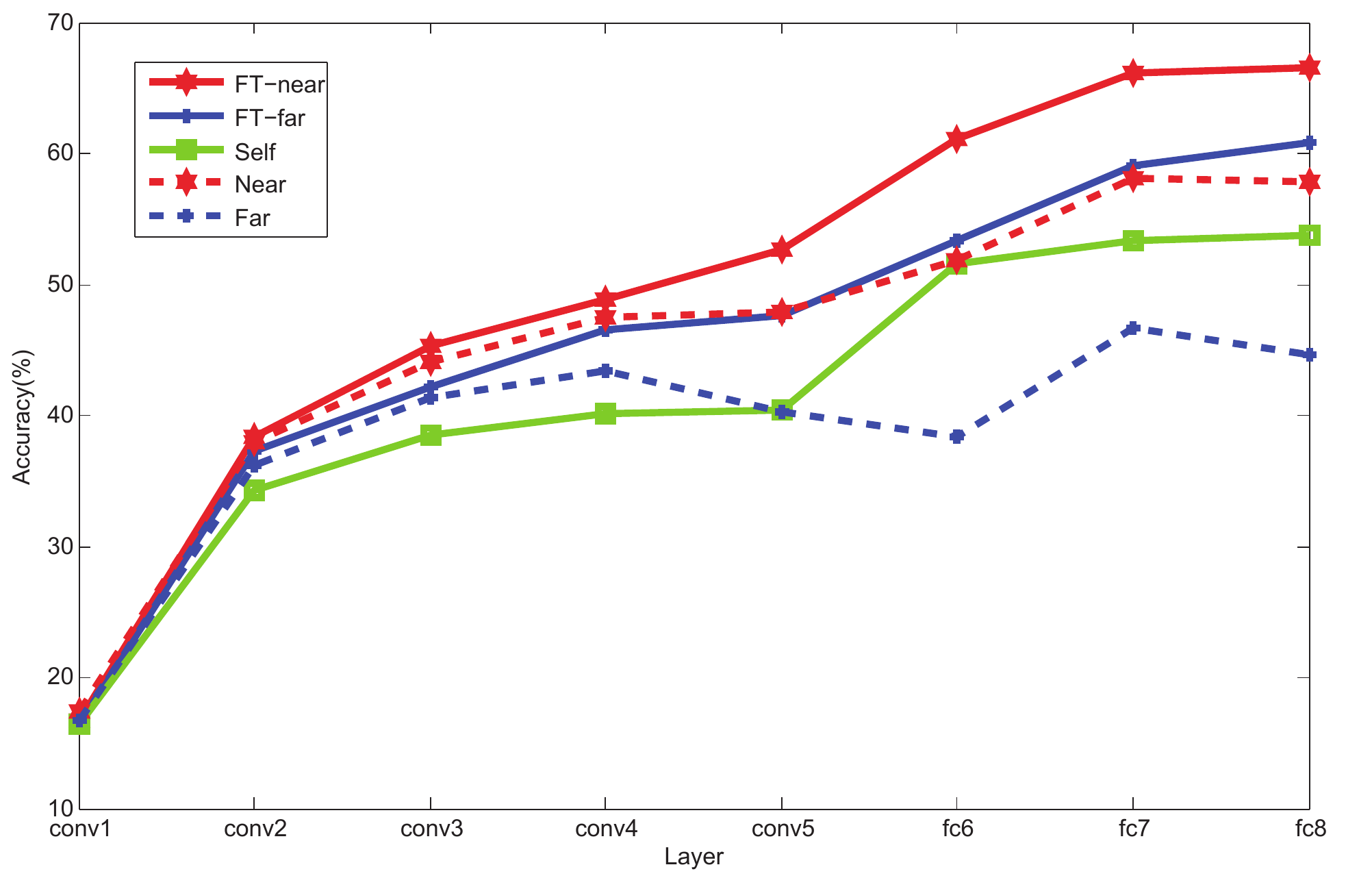}}\tabularnewline
\multicolumn{2}{c}{(b) The gain of the the performance}\tabularnewline
\includegraphics[width=0.48\textwidth]{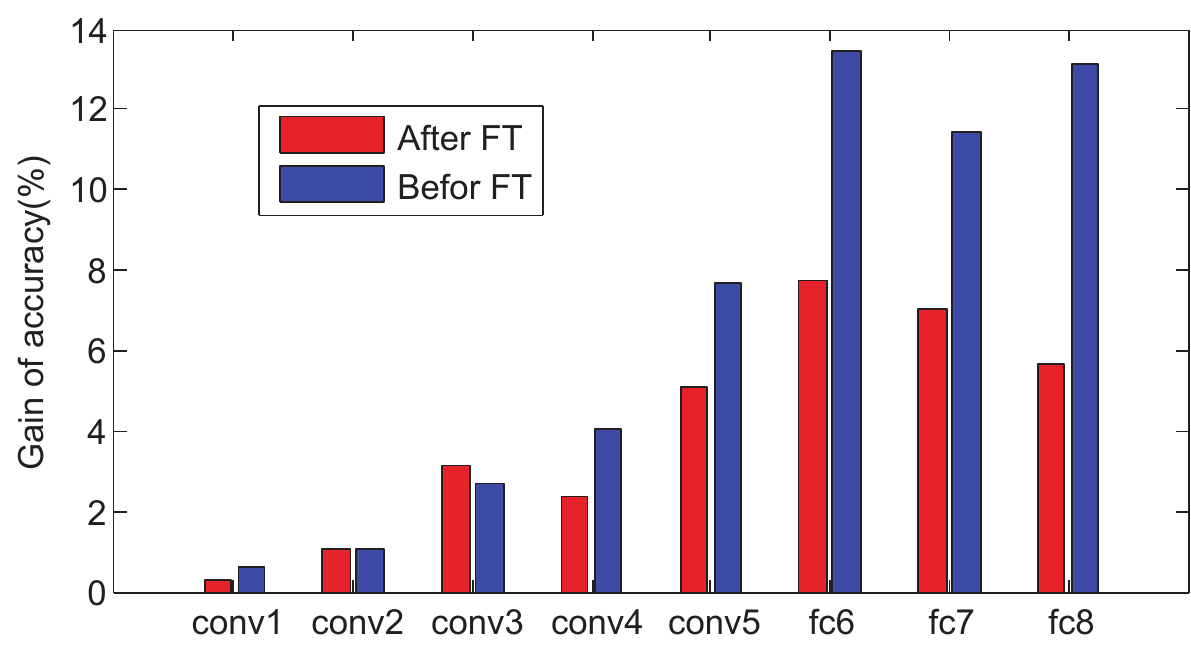} & \includegraphics[width=0.48\textwidth]{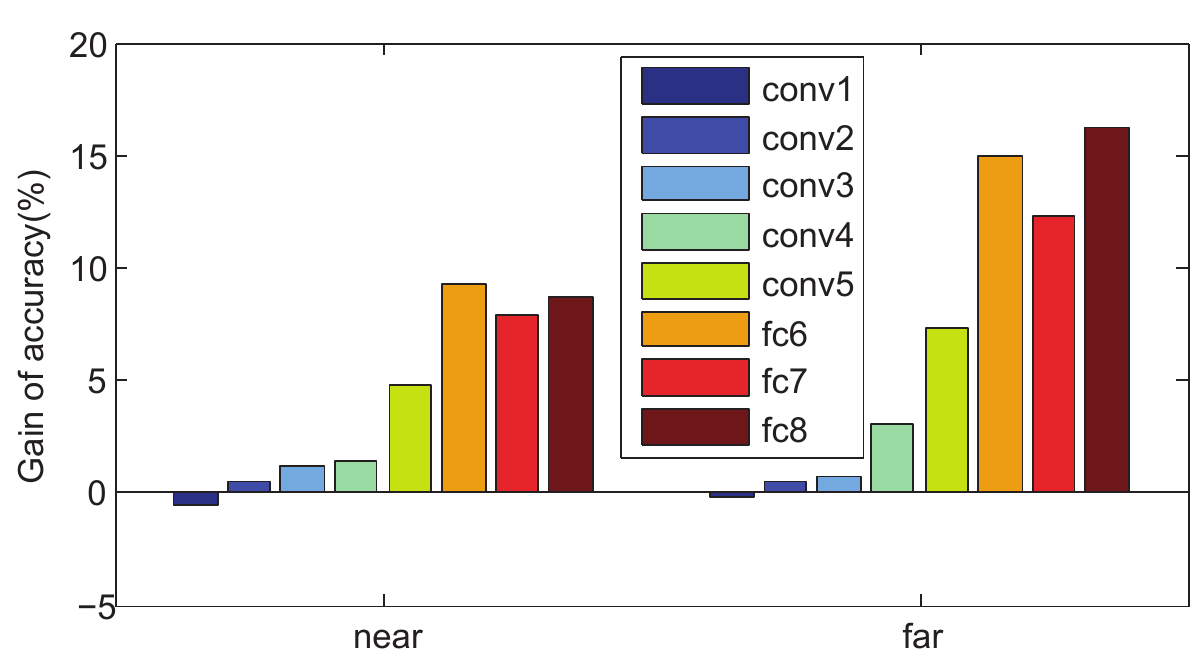}\tabularnewline
\end{tabular}
\par\end{center}%
\end{minipage}

\caption{The performance on INSTRU. (a) The discrimination of different layers
when the features extracted from different models.\label{fig:The-discrimination-of-different-layer}
Self represents the features extracted from the model which is trained
on the 100 categories dataset of INSTRU. Near and Far represent the
model trained on the 500 categories in dataset near\_INSTRU and far\_INSTRU
respectively. FT-near indicates fine-tuning the model pre-trained
on near\_INSTRU on the target dataset, so does FT-far. (b) The left
illustrates gain of the performance that the near model surpasses
than the far one for each layer.\textcolor{black}{{} The right illustrates
gain of the performance that after fine-tuning surpasses before fine-tuning
for the near and far models respectively.}}
\end{figure*}

\subsubsection{The discrimination of different layers\label{subsec:The-discrimination-of-layers}}

\textcolor{black}{It is valuable to measure the discrimination of
different layer features extracted from different models, as low layer
features are common for all images and high layer features are specific
to the task}\textcolor{red}{\ }\textcolor{black}{\cite{VISUAL_2014}.}
We analyze the classification accuracy of features extracted from
layer conv1 to layer fc8 on different models on the dataset of INSTRU.
Here conv1 represents the activation of the first convolutional layer
followed with the Relu, Pooling and Norm, with a dimension of $96\times27\times27$.
Conv2 and conv5 are the same situations.\textcolor{black}{{} We use
a logistic regression classifier composed with a fully-connected layer
of 100 units and a softmax layer with all training examples exposed
to the training procedure. The results are shown in Fig.~\ref{fig:The-discrimination-of-different-layer}.
}Many novel conclusions can be obtained. 

\textcolor{black}{First, the generalization capability of the activations
of a pre-trained CNN model increases with the growth of the layer.
What is more, if we transfer the features from a model which is trained
on a source dataset which is different with the one we evaluate the
performance (i.e., the target dataset), the activations of fc7 are
the best. But if we train or fine-tune the model on the dataset itself,
the activations of fc8 (which have smaller dimensions) are the best.
}As shown in the solid lines of Fig.~\ref{fig:The-discrimination-of-different-layer}
(a), the features of fc8 (66.62\%, 60.86\%, 53.74 respectively) obtain
higher performance compared with the features of fc7 (66.12\%, 59.08\%,
53.32\% respectively). However, the features of fc8 (57.78\%, 44.64\%
respectively) obtain lower performance in all dash lines. 

Second, for models trained on datasets that have different distances
to target dataset, the performance of the near one is much higher
than the performance of the far one. The case is always true for all
the layers. What is more, fine-tuning from these models, the trend
is still remained. \textcolor{black}{This is corresponded with the
conclusion that the more similar between the source and target datasets,
the better performance fine-tuning obtains.}

\textcolor{black}{Third, the gain of the performance that the near
model surpasses than the far one decreases after fine-tuning, as shown
in the left of Fig.~\ref{fig:The-discrimination-of-different-layer}
(b).} The filters optimized on the dataset which is near to the target
dataset are sensitive to the images in the target dataset, while filters
optimized on the far dataset seem insensitive, so the difference between
these two sets of filters is big. After fine-tuning on the target
dataset, filters in these models both fire on the target images, so
the difference decreases.

\textcolor{black}{Fourth, we compare the performance of pre-trained
models with the one trained from scratch.} Examples in the near source
dataset have many common attributes or high level visual patterns
with the ones in the target dataset also have. Meanwhile, the source
dataset has bigger training examples ($500\times1300=650000$) than
the training examples in the target dataset ($100\times1300=130000$),
so representations extracted from the model trained on the near source
dataset (Near) are better than the features extracted from the model
trained on the target dataset itself (Self). However, examples in
the far source dataset have less common high level visual patterns,
so features extracted from the far model (Far) are inferior to the
features extracted from Self.\textcolor{red}{{} }\textcolor{black}{However,
the performance of Far in the layers bellow conv5 is better than the
performance of Self, as illustrated in Fig.~\ref{fig:The-discrimination-of-different-layer}
(a). }The reason is that features in the low layers are common patterns
for general object recognition, which have little relationship with
the specific target task. With more training examples fed to Far,
the model captures more simple visual patterns, leading a better performance.

\subsubsection{Qualitatively analysis}

\begin{figure*}
\ \ \ \ \ %
\noindent\begin{minipage}[t]{1\columnwidth}%
\setlength{\tabcolsep}{2pt}
\begin{center}
\begin{tabular}{cc}
\includegraphics[width=0.46\textwidth]{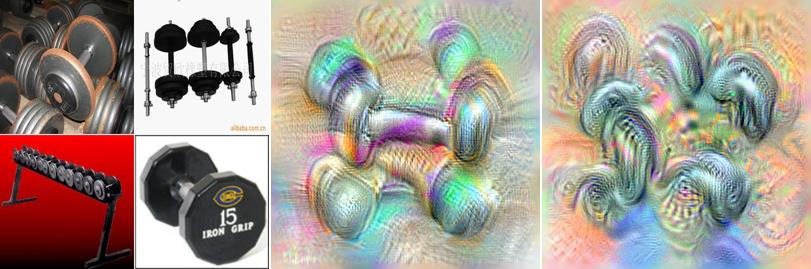} & \includegraphics[width=0.46\textwidth]{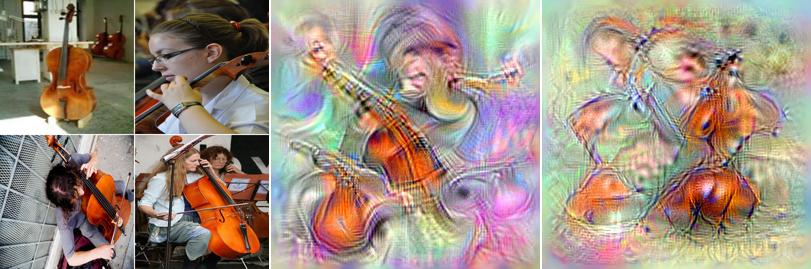}\tabularnewline
\includegraphics[width=0.46\textwidth]{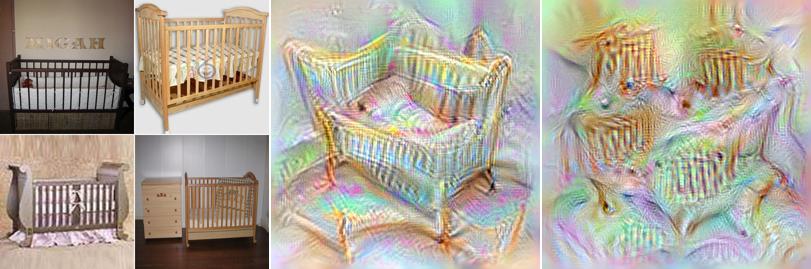} & \includegraphics[width=0.46\textwidth]{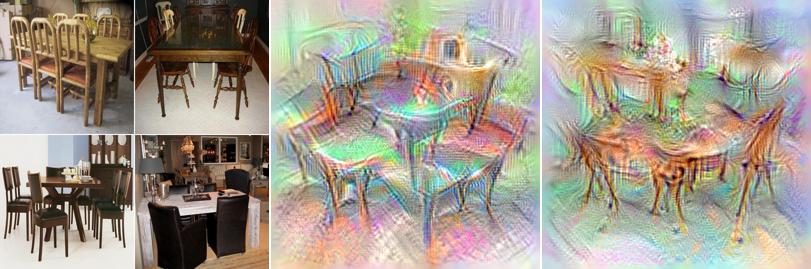}\tabularnewline
\includegraphics[width=0.46\textwidth]{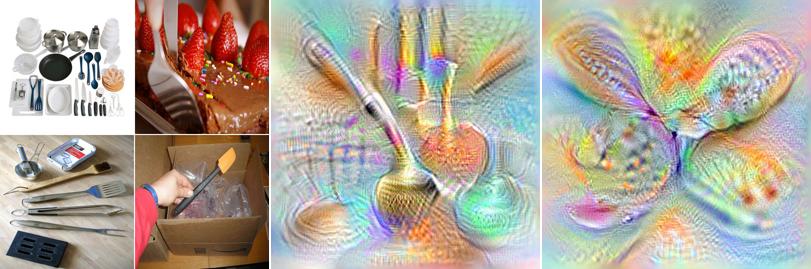} & \includegraphics[width=0.46\textwidth]{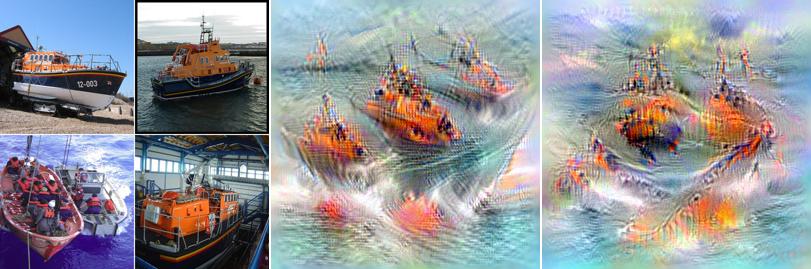}\tabularnewline
\end{tabular}
\par\end{center}%
\end{minipage}

\caption{The visualization of class images for different models.\label{fig:Qualitatively-analysis-of}
For each set, the first column shows same original images, and the
second column is the corresponding class image computed from the model
fine-tuned on near\_INSTRU (FT-near). The third column is the corresponding
image computed from the model fine-tuned on far\_INSTRU (FT-far).
The results show that class images computed from FT-near are better
in representing the corresponding classes than the class images computed
from FT-far (Best viewed in color).}
\end{figure*}

\textcolor{black}{We also qualitatively analyze the difference between
different models. For a learned model, we utilize the method introduced
by \cite{VDEEO_INSIDE_2014} to visualize the class images. More formally,
let $S_{c}(I)$ be the score of the class $c$, which is computed
by the classification layer of the model for image $I$. }We would
like to find the image $I_{c}$, such that the score $S_{c}$ is high.
So $I_{c}$ can be regarded as the the notion of class $c.$ Fig.
\ref{fig:Qualitatively-analysis-of} qualitatively shows some examples. 

We can see that the fine-tuned near model (FT-near) learns more details
of the classes in INSTRU, compared with the fine-tuned far model (FT-far).
For example, as shown in the first set Fig.\ \ref{fig:Qualitatively-analysis-of},
FT-far only focuses on the balls of a\emph{ dumbbell}. In comparison,
FT-near not only focuses on the balls but also the short bar which
serves as a handle. All of the examples show that the class images
computed from FT-near are more representative than the ones computed
from FT-far. \textcolor{black}{The results demonstrate that the learned
features in FT-near are better in representating images in INSTRU.}

\begin{figure*}
\begin{centering}
\begin{tabular}{c}
\includegraphics[width=0.88\textwidth]{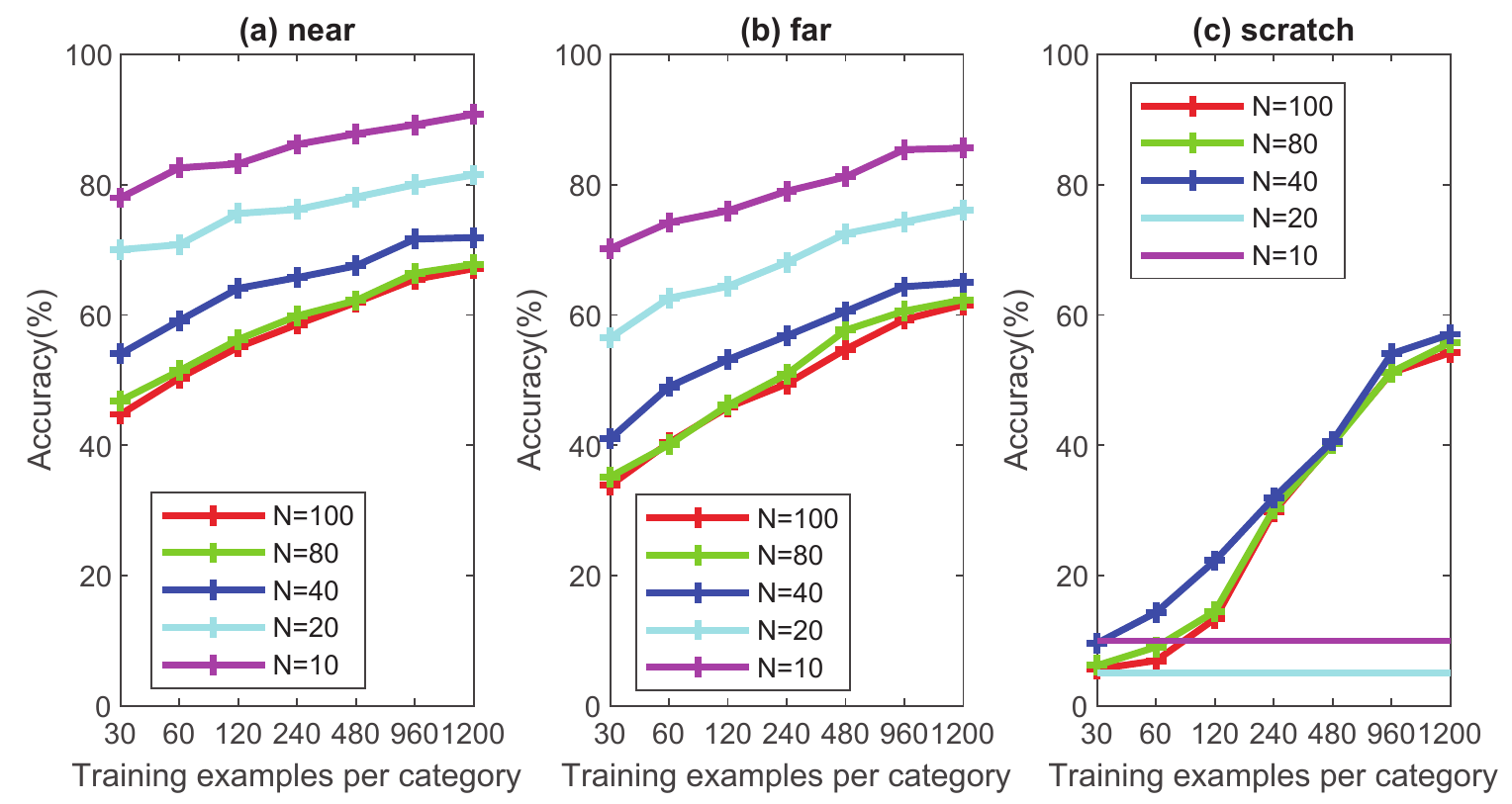}\tabularnewline
\includegraphics[width=0.68\textwidth]{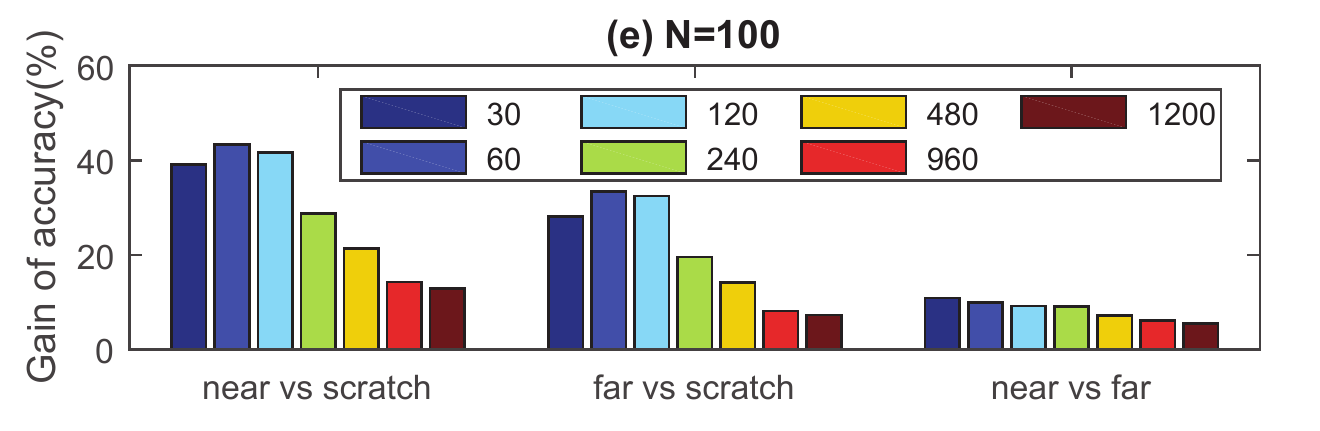}\tabularnewline
\includegraphics[width=0.66\textwidth]{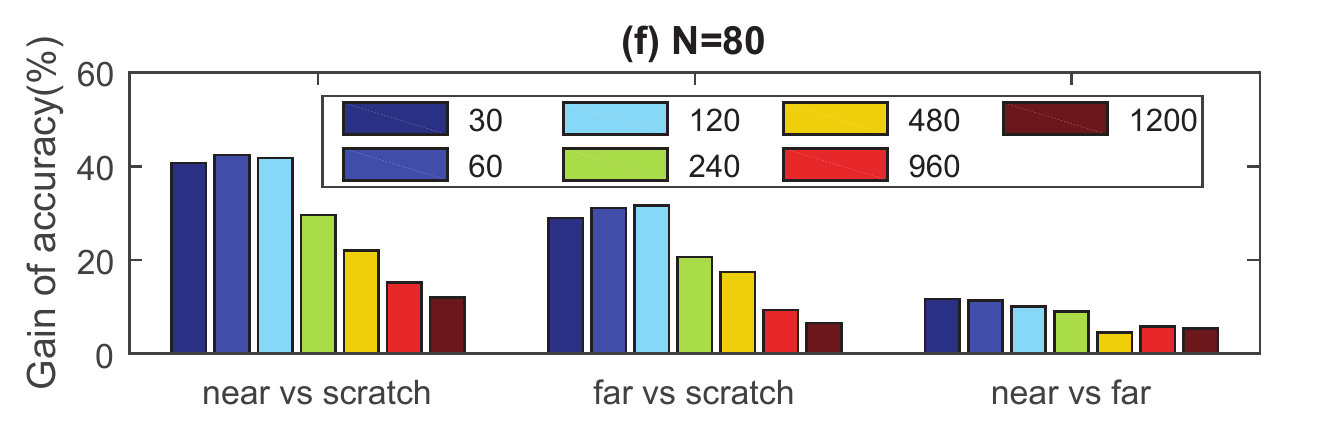}\tabularnewline
\end{tabular}
\par\end{centering}
\caption{\textcolor{black}{The effect of number of categories and training
samples per category in the target dataset on the performance of fine-tuning.}\label{fig:The-effect-of-per-number-1}
The horizontal lines in (c) represent that the models do not converge.
(e) shows the comparison of two models when the number of categories
in the target dataset is 100 ($N=100$). For example, \textquotedblleft near
vs scratch\textquotedblright{} shows the gains of fine-tuning the
model which is pre-trained on near\_INSTRU (near) versus training
the model from scratch (scratch) with the training samples per category
varying from 1200 to 30. (f) shows the results under the settings
of $N=80$.}
\end{figure*}

\subsection{The Number of Categories and Training Examples per Category in the
Target Dataset}

\textcolor{black}{The number of categories and training samples per
category play an important role in training a convolutinal neural
network. In this section, we analyze the effect of these two factors
in the target dataset when conducting fine-tuning. }As the average
training examples for each category in ImageNet are almost 1200, we
fine-tune the pre-trained models on target datasets with the training
samples per category varying from 1200 to 30. At the same time, we
also select 100, 80, 40, 20, 10 as the number of categories in the
target datasets respectively. We use the models trained on far\_INSTRU
and near\_INSTRU (introduced in Section\ \ref{subsec:The-distance-between})
as two pre-trained models and select target datsets with different
settings from INSTRU.

As illustrated in Fig.~\ref{fig:The-effect-of-per-number-1}, the
results suggest that with the decrease of the categories in the target
datasets, the performance of fine-tuning from a pre-trained increases\textcolor{black}{.
It is true both with the far and near models as well as training from
scratch.} With the number of categories decrease, the accuracy may
decrease as less training examples are fed to the network, which could
more likely tend to overfiting. But the results illustrate that the
accuracy is higher, because the the classification problem goes easier
with less categories, which has less chances to make mistakes. It
is important to note that when the data is too scarce to train a model,
as shown in Fig. \ref{fig:The-effect-of-per-number-1} (c), fine-tuning
a pre-trained model on such a small dataset can also get significant
performance, but training from scratch does not converge. The second
conclusion we can obtain is that when the number of categories in
the target dataset is fixed, the accuracy gets higher when more training
examples are exposed to the model. The results show that more training
examples can give better performance.\textcolor{red}{{} }\textcolor{black}{The
third conclusion is that the gain (i.e., near vs scratch, far vs scratch)
decreases with the increase of training examples. The trend is the
same for the gain of near versus far, as illustrated in Fig.\ \ref{fig:The-effect-of-per-number-1}\ (e)
and (f). With more training examples, the parameters in different
models both are adjusted to fit the target data, being more sensitive
to the images in the target dataset, so the difference decreases.}

\subsection{Fine-tuning to Deep or Shallow Layers}

Features in low layers of convolutional neural networks are general
image representations such as common edges, shapes and textures, while
features in high layers are specific image representations for the
target task. So when we fine-tune a pre-trained model on the target
dataset, the parameters of the model in which layer should be fixed,
and which layer should be retrained?\textcolor{black}{{} The work of
Yosinki~\cite{TRANS_FT} has suggested that fine-tuning all the layers
can obtain significant performance, while fixed the parameters of
some layers will bring about the problems of co-adaptation (for middle
layers) and representation specificity (for high layers). }Their work
provides a sufficient analysis of fine-tuning when the training examples
are abundant. But what are the results when the training examples
are limited? In this section, we investigate the problem of fine-tuning
to deep or shallow layers in a systematic way on the condition of
limited training data. 

\begin{figure}
\noindent\begin{minipage}[t]{1\columnwidth}%
\begin{center}
\includegraphics[width=0.7\textwidth]{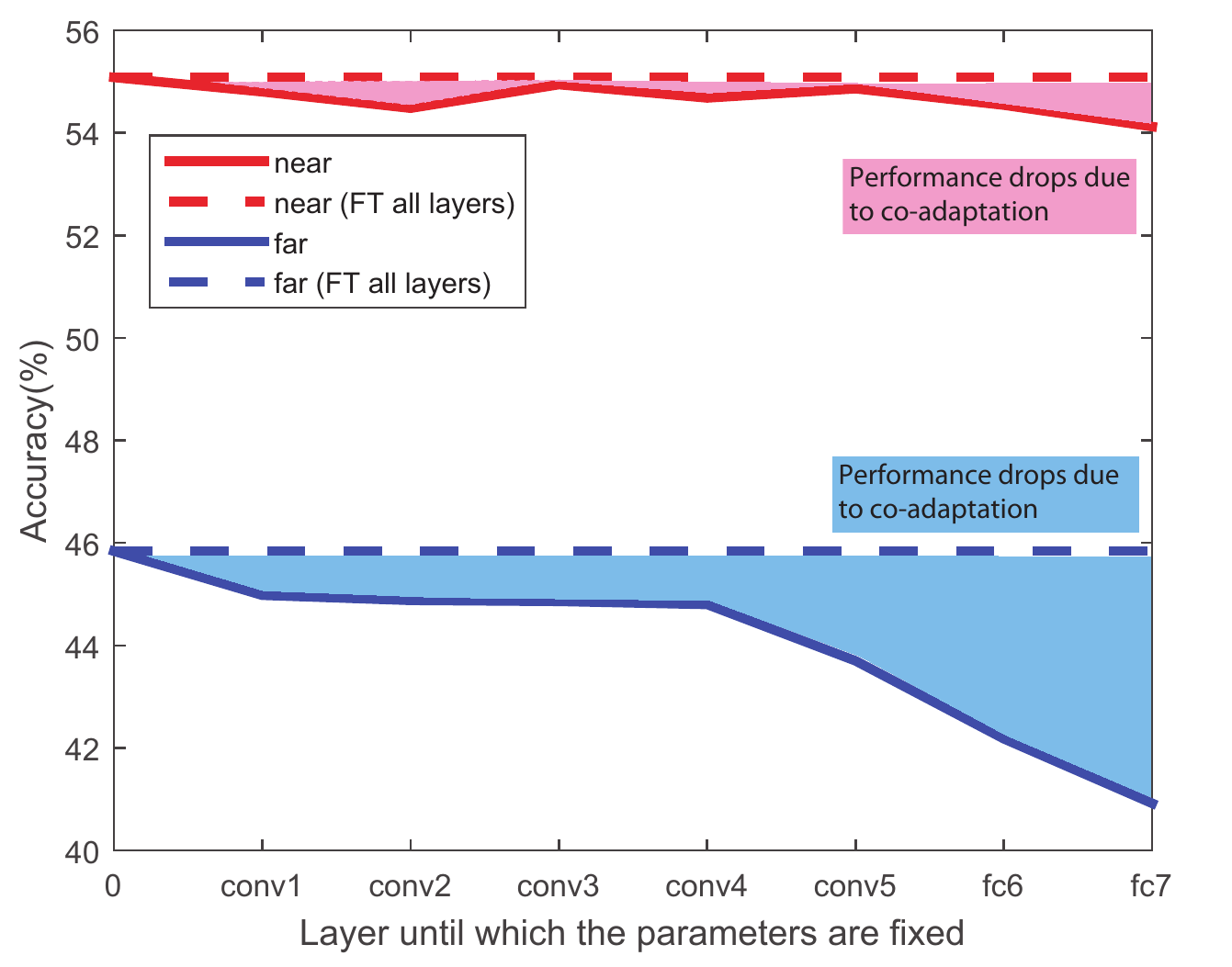}
\par\end{center}%
\end{minipage}

\caption{The performance of fine-tuning to deep or shallow layers on different
models when the training data is limited.\label{fig:The-performance-of-u-layer}
Red and blue lines represent fine-tuning from the near model and the
far model respectively.}
\end{figure}

In this section, we also use the models trained on far\_INSTRU and
near\_INSTRU (introduced in Section\ \ref{subsec:The-distance-between})
as two pre-trained models (denoted as near and far respectively).
For the target dataset, we use all the categories in INSTRU, but only
sample 240 examples per category for training (the original training
examples per category is about 1200). In the experiments, the first
$n$ (from 1 to 7) layers of the network are frozen. For example,
when $n=3$, the parameters for the layers conv1, conv2 and conv3
are fixed during fine-tuning. The results are shown in Fig.~\ref{fig:The-performance-of-u-layer}.
The first conclusion we can get is that it is better to fine-tune
all the layers on the target dataset. This is consistent with the
conclusion in \cite{TRANS_FT} where the training examples in the
target dataset is large. \textcolor{black}{The second conclusion is
that there is less co-adaptation (i.e., features that interact with
each other in a complex or fragile way such that this co-adaptation
could not be relearned by the upper layers alone\ \cite{TRANS_FT})
in the model which is fine-tuned from near, compared to the model
which is fine-tuned from far. Because the parameters in near is already
fitted to the target data very well. They can be easily transferred
to classifying the target data. It also can be concluded that just
fine-tuning the high layers such as conv5, fc6 and fc7 can yield a
satisfactory results, when the training data is small.}

\subsection{The Effect of Data Augmentation}

Data augmentation is a common and simple approach to learning discriminative
and representative features by augmenting the training set with transformed
versions of the original images. Techniques such as such as cropping,
rotating, scaling, and flipping original input images have been widely
used for image classification. In this section, we evaluate the effect
of data augmentation for fine-tuning. We artificially constrain our
access to the target training data, and evaluate the gain in recognition
accuracy which is brought by data augmentation.

\begin{figure}
\begin{centering}
\includegraphics[width=0.78\textwidth]{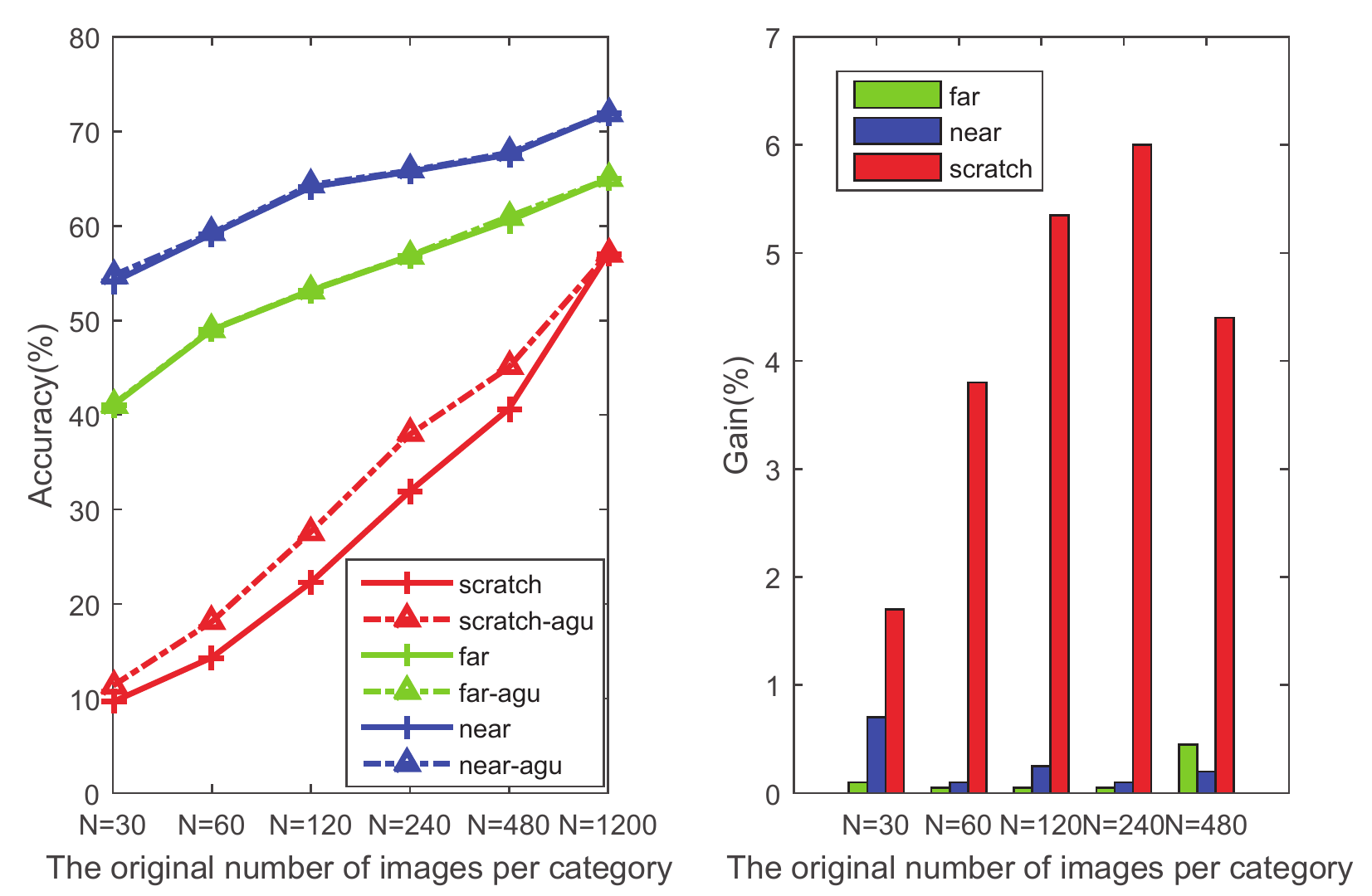}
\par\end{centering}
\caption{\textcolor{black}{The comparison of data augmentation for fine-tuning
and training models from scratch.\label{tab:The-effect-of-data_augmentation}
For original number of images that is smaller than 1200, we increase
the training examples to 1200 by data augmentation. For the left figure,
red lines represent training network from scratch. Blue and green
lines represent fine-tuning network from the near model and the far
model respectively. The right figure shows the gain which is brought
by data augmentation (Best viewed in color).}}
\end{figure}

\textcolor{black}{For the target datasets, we use all the categories
in INSTRU, with the original training samples per category varying
from 1200 to 30. Two pre-trained models (introduced in Section\ \ref{subsec:The-distance-between},
denoted as near and far) are fine-tuned on these datasets respectively.
}As the original training examples per category is about 1200, we
set 1200 as the upper bound. In our experiments, if the original number
of images per category in the training set is smaller than 1200, we
use the data augmentation technique in~\cite{TPI} to augment the
training images to 1200.

The results are shown in Fig.~\ref{tab:The-effect-of-data_augmentation}.
We compare the gain in accuracy when training network from scratch,
fine-tuning network from the near model and the far model respectively.
The conclusion is that data augmentation brings significant gains
for training neural networks from scratch. While for fine-tuning,
the gains are smaller. The performance for fine-tuning with and without
data augmentation is almost the same.\textcolor{red}{{} }\textcolor{black}{For
training models from scratch, as models trained with small data do
not generalize well, data augmentation increases the amount of training
data, enhancing the ability to correctly classify images. However,
a pre-trained model is initialized with proper weights. Fine-tuning
it on the generated examples brings small changes. So data augmentation
for fine-tuning is less effective than data augmentation for training
networks from scratch.}

\section{Conclusion\label{sec:Conclusion}}

\textcolor{black}{Features extracted from CNNs trained on large scale
datassets are becoming common and preferred image representations
for more and more visual tasks. Fine-tuning on a pre-trained model
not only saves lots of time to avoid training networks from scratch,
but also guarantees better performance. Our paper is dedicated to
analyzing the factors that influence the performance of fine-tuning.
We constrain our access to the subsets extracted from ImageNet, and
explore each factor in turn. Many observations can be concluded from
the results. First, with roughly the same distance to the target dataset,
the bigger of the number of categories in the source dataset, the
better performance fine-tuning obtains. Second, when the source dataset
is fixed, the performance of fine-tuning increases with more training
examples exposed to the retraining of the pre-trained model. What
is more, the gain of fine-tuning versus training network from scratch
decreases with the increase of retraining examples. Third, the more
similar between the source and target datasets, the better performance
fine-tuning obtains. We analyze the characteristic of different models
at both filter-level and layer-level, and show their sensitivities
to the target dataset. These conclusions provide useful and evidence-backed
intuitions about how to implement fine-tuning for other visual tasks.
In future work, we will explore architectures with an auxiliary loss
to leverage the relationship between the source and target datasets
explicitly when conducting fine-tuning.}

\bibliographystyle{ACM-Reference-Format}
\bibliography{main-ft}

\end{document}